\documentclass[12pt]{article}
\usepackage{amsmath}
\usepackage{graphicx}
\usepackage{float} 
\usepackage{authblk}
\usepackage[numbers]{natbib}
\usepackage{hyperref}
\usepackage{natbib}
\usepackage{authblk}
\usepackage{amsmath,amsfonts,amssymb}
\usepackage{caption}
\usepackage{subfig}  
\usepackage{subcaption}
\usepackage{url}
\usepackage{xcolor}
\definecolor{mygreen}{RGB}{0,128,0}

\title{A physics informed neural network approach to simulating ice dynamics governed by the shallow ice approximation}

\author[1]{Kapil Chawla\thanks{Email: kapchaw@iu.edu}}
\author[2]{William Holmes\thanks{Email: wrholmes@iu.edu}}
\affil[1]{Department of Mathematics and the Institute for Scientific Computing and Applied Mathematics, Indiana University, Indiana 47405, USA}
\affil[2]{Department of Mathematics and Cognitive Science Program, Indiana University, Indiana 47405, USA}
\affil[*]{Corresponding author: Kapil Chawla, kapchaw@iu.edu}

\begin{document}

\maketitle

\begin{abstract} 
In this article we develop a Physics Informed Neural Network (PINN) approach to simulate ice sheet dynamics governed by the Shallow Ice Approximation. This problem takes the form of a time-dependent parabolic obstacle problem \cite{PT2023}. Prior work has used this approach to address the stationary obstacle problem and here we extend it to the time dependent problem \cite{KC2024}. Through comprehensive 1D and 2D simulations, we validate the model’s effectiveness in capturing complex free-boundary conditions. By merging traditional mathematical modeling with cutting-edge deep learning methods, this approach provides a scalable and robust solution for predicting temporal variations in ice thickness. To illustrate this approach in a real world setting, we simulate the dynamics of the Devon Ice Cap, incorporating aerogeophysical data from 2000 \cite{Dowdeswell2004} and 2018 \cite{Rutishauser2022}.

\end{abstract}

\section{Introduction}

Grounded ice thickness plays a critical role in understanding the behavior and stability of ice sheets, particularly in polar regions such as Greenland, Antarctica, and the Canadian Arctic. Ice sheet dynamics are governed by complex interactions between ice flow, surface accumulation, and bedrock topography, making the accurate modeling of these processes essential for predicting long-term ice sheet behavior and their contributions to global sea level rise \cite{Ed2012, PT2023}. In particular, the Shallow Ice Approximation (SIA) provides a framework for modeling grounded ice, where ice flow is driven by internal deformation and the base is often assumed to be frozen, constraining the ice thickness by bedrock topography \cite{R2009, jouvet_rappaz_bueler_blatter_2011}.

A key challenge in modeling grounded ice involves solving the partial differential equations (PDEs) that govern ice thickness evolution, while incorporating these constraints. This leads to a free boundary problem, where the ice thickness must remain non-negative and above the bedrock, giving rise to an obstacle problem \cite{JF1987, calvo2002}. In previous work, we addressed the stationary obstacle problem using a neural network approach to solve for steady-state ice thickness \cite{KC2024,KC2024arxiv}. In this paper, we extend that framework to handle the time-dependent evolution of ice thickness, framing the grounded ice thickness problem as a parabolic obstacle problem.

In this work, we develop a machine learning framework to solve the grounded ice thickness problem using a Physics-Informed Neural Network (PINN). By framing the grounded ice problem as a parabolic obstacle problem, we leverage PINNs to approximate the solution of the ice thickness equation while ensuring that the obstacle and boundary conditions are satisfied \cite{Raissi2019, Rudy2019}. The neural network minimizes a composite loss function that includes terms for the PDE residual, obstacle constraint, and boundary conditions, providing a flexible and scalable method for solving this problem \cite{Khoo2019, Cheng2023}.

We apply this framework to the Devon Ice Cap in the Canadian Arctic, utilizing ice thickness data from aerogeophysical surveys conducted in 2000 and 2018. By training the model on the 2000 data and validating it against the 2018 data, we demonstrate the model's ability to accurately predict the evolution of grounded ice thickness over time \cite{Bamber2001, KC2024}.

This paper is organized as follows. Section 2 presents the time-dependent parabolic obstacle problem and the penalty method used to solve it \cite{R1984}. This simplified problem shares structural similarities with the ice sheet model that is our ultimate goal and provides a more accessible setting to develop a computational solution approach. Section 3 describes the approximation of the solution using a Physics-Informed Neural Network (PINN) approach, including numerical results, optimization and training processes, and examples in both 1D and 2D. Section 4 introduces the mathematical model for ice sheet dynamics \cite{R1994} and discusses the singularity, including that in the time derivative term. Section 5 presents numerical computations for the ice sheet model in 1D and 2D using manufactured solutions. Section 6 applies the developed techniques to the Devon Ice Cap data, demonstrating the model's effectiveness in real-world scenarios. Finally, Section 7 concludes the paper with key insights and outlines directions for future research.

\section{Solving the Parabolic Obstacle Problem Using the Penalty Method}

Here we formulate a simplified parabolic obstacle problem. This is similar in structure to the ice sheet model we consider in later sections. However it is simplified and thus provides a good beginning test bed to assess our solution method. Below we describe the variational formulation of this problem, the associated penalty method formulation, and the loss terms we will use in the formulation of our PINN problem.

\subsection{Problem Setup}

We consider the space-time domain \( Q = \Omega \times (0, T) \), where \( \Omega \subset \mathbb{R}^n \) is a bounded domain with smooth boundary \( \partial \Omega \) (for this paper $n=1,2$). The goal is to find a solution \( u(t,x) \) such that:

\begin{equation}
\left\{
\begin{aligned}
    & \text{Find } u(t, x) \in K(\psi) \text{ such that:} \\
    & \int_Q \left( u_t (v - u) + \nabla u \cdot \nabla (v - u) \right) dx \, dt \geq \int_Q f (v - u) \, dx \, dt, \\
    & u \geq \psi \quad \text{in} \; Q, \\
    & u = g \quad \text{on} \; \partial \Omega \times (0, T), \\
    & u(0,x) = u_0(x) \quad \text{in} \; \Omega.
\end{aligned}
\right.
\end{equation}

\noindent for all \( v \in K(\psi) \), where \( K(\psi) = \{ v \in H_0^1(\Omega): v \geq \psi \} \). The obstacle \( \psi \), boundary condition \( g \), and initial condition \( u_0 \) are given, and \( f \) represents a source term.

\subsection{Penalty Method Formulation}

To solve the variational inequality, we approximate the obstacle constraint using a penalization approach. The penalized form of the problem seeks a weak solution to the following system of equations:

\[
u^\epsilon_t - \Delta u^\epsilon - \frac{1}{\epsilon} \max(0, \psi - u^\epsilon) = f \quad \text{in} \; Q,
\]
where \( \epsilon > 0 \) is the penalty parameter. As \( \epsilon \to 0 \), the penalized problem approaches the original obstacle problem \cite{Geshkovski2024}.

\subsection{Solution Approximation Using Deep Neural Networks}

Here we develop a PINN based approach to compute solutions to this time dependent obstacle problem with spatial dimensions $n=1,2$. We employ a fully connected feedforward neural network, denoted as \( \mathbf{f}: \mathbb{R}^d \rightarrow \mathbb{R} \), to approximate the solution of the parabolic obstacle problem. Here the domain of $\mathbf{f}$ is the time-space domain and its output is the elevation at that time and place: $\mathbf{f} : (t,x) \rightarrow u(t,x)$. The network parameters (weights and biases) are iteratively optimized through backpropagation to minimize a composite loss function that enforces the solution's adherence to the PDE, obstacle constraints, and boundary conditions.

To guide the neural network toward an accurate solution, we define a total loss function composed of several key components. In the below, $(t_i,x_i)$ are co-location points in either the interior of the domain or its boundary. 

\begin{enumerate}
    \item \textbf{PDE Residual Loss}: This loss term ensures the solution satisfies the underlying PDE. Here the colocation points are in the interior of the domain.
    \[
    L_{\text{PDE}}(u) = \frac{1}{N_{\text{PDE}}} \sum_{i=1}^{N_{\text{PDE}}} \left( u_t(t_i, x_i) - \Delta u(t_i, x_i) - f(t_i, x_i) \right)^2.
    \]

    \item \textbf{Obstacle Constraint Loss}: This term penalizes violation of the obstacle condition. Smaller values of $\epsilon$ more strognly enforce the obstacle constraint. The colocation points here are in the interior of the domain.
    \[
    L_{\text{penalty}}(u) = \frac{1}{\epsilon N_{\text{obs}}} \sum_{i=1}^{N_{\text{obs}}} \max\left(0, \psi(t_i, x_i) - u(t_i, x_i)\right)^2.
    \]
    
    \item \textbf{Boundary Condition Loss}: This term ensures the solution satisfies the boundary conditions. The colocation points here are on the domain boundary.
    \[
    L_{\text{boundary}}(u) = \frac{1}{N_{\text{boundary}}} \sum_{i=1}^{N_{\text{boundary}}} \left( u(t_i, x_i) - g(t_i, x_i) \right)^2.
    \]

    \item \textbf{Initial Condition Loss}: This term ensures the solution satisfies the initial condition.
    \[
    L_{\text{initial}}(u) = \frac{1}{N_{\text{init}}} \sum_{i=1}^{N_{\text{init}}} \left( u(0, x_i) - u_0(x_i) \right)^2.
    \]
\end{enumerate}

\noindent The total loss function is then defined as:

\[
L_{\text{total}}(u) = L_{\text{PDE}}(u) + L_{\text{penalty}}(u) + L_{\text{boundary}}(u) + L_{\text{initial}}(u).
\]

This composite loss function ensures that the neural network approximates the solution to the PDE, while respecting the obstacle condition and satisfying both the initial and boundary conditions. The neural network is trained using the Adam optimizer, with a learning rate scheduler to adjust the learning rate dynamically during training. By iteratively minimizing the total loss function, the network parameters are refined to produce an accurate approximation of the solution. The specifics of the computational domain and the colocation points chosen within it are discussed in the 1D and 2D problem sections below.

\subsection{Numerical Results}
Before delving into the specific case studies, we first describe the numerical experiments and methods utilized in this research. The main objective is to approximate solutions to parabolic obstacle problems using a deep learning framework. To achieve this, we employ a fully connected neural network architecture designed to handle both 1D and 2D parabolic obstacle problems.

We begin by developing a manufactured solution to assess the precision of our method. This approach, called the \textit{Method of Manufactured Solutions (MMS)}, is frequently employed in computational science to evaluate numerical techniques by generating an exact solution for a modified form of the PDE. In our case, we select a solution that satisfies both the obstacle and boundary conditions. The residuals, which result from this solution not fulfilling the original PDE, are used to adjust the PDE with an additional inhomogeneous term. The exact solution to this modified PDE serves as a reference against which the neural network’s approximation is compared. By using MMS in these initial experiments, we can thoroughly validate the reliability and accuracy of the deep learning method.

After verifying the approach through MMS, we move on to solving the parabolic obstacle problem directly. This problem involves a PDE with a challenging obstacle constraint. To fully evaluate the performance of the method, we first address a simpler PDE with an obstacle constraint, then tackle the more complex parabolic PDE of interest. These computations are performed on both 1D and 2D domains. Additionally, we apply the method to a practical problem by estimating the elevation profile of ice above the bedrock of the Devon Ice Cap.

For the neural network architecture, we utilize a fully connected design that includes at least five hidden layers, each containing 128 neurons. The \textit{squared Rectified Linear Unit (ReLU\(^2\))} is chosen as the activation function due to its superior differentiability and strong empirical performance when addressing complex Partial Differential Equations (PDEs). This activation function enables the network to better capture nonlinear behaviors in the solution, promoting smoother and more efficient optimization \cite{Jagtap2020adaptive}.

\subsection{Optimization and Training}

For training of this solution neural network solution approximator, we use the Adam variant of stochastic gradient descent (SGD). The learning rate is initially set to \( 5 \times 10^{-4} \) for the first 500 iterations. After that, the learning rate is reduced by half between iterations 500 and 750, and further halved for the remaining iterations. This learning rate schedule was determined through trial and error, ensuring stable convergence of the neural network. The primary objective throughout the optimization process is the minimization of the loss functions.

For the implementation and training of our model, we rely on the \textit{PyTorch} library, which provides the necessary tools for automatic differentiation and GPU acceleration. PyTorch’s flexibility allows us to efficiently solve the parabolic obstacle problems and adjust the network architecture as needed. 

\subsection{1D Example}

We consider a 1D parabolic obstacle problem in the spatial domain \( \Omega = [0,1] \) and temporal domain \( t \in [0,1] \), where both the obstacle function \( \psi(x,t) \) and the exact solution \( u(x,t) \) are piecewise-defined functions of space and time. The obstacle function is defined as:

\[
\psi(x, t) = 
\begin{cases} 
100 x^2, & 0 \leq x < 0.25, \\
100 x(1 - x) - 12.5, & 0.25 \leq x < 0.5, \\
100 x(1 - x) - 12.5, & 0.5 \leq x < 0.75, \\
100 (1 - x)^2, & 0.75 \leq x \leq 1.
\end{cases}
\]

\noindent The exact solution to the problem is given by the following piecewise-defined function:

\[
u(x, t) = 
\begin{cases} 
(100 - 50\sqrt{2}) x \exp(-\gamma t), & 0 \leq x < \frac{1}{2\sqrt{2}}, \\
100 x(1 - x) - 12.5 \exp(-\gamma t), & \frac{1}{2\sqrt{2}} \leq x < 0.5, \\
100 x(1 - x) - 12.5 \exp(-\gamma t), & 0.5 \leq x < 1 - \frac{1}{2\sqrt{2}}, \\
(100 - 50\sqrt{2}) (1 - x) \exp(-\gamma t), & 1 - \frac{1}{2\sqrt{2}} \leq x \leq 1.
\end{cases}
\]

\noindent In the following experiments, we set \( \gamma = 0.001 \), which controls the rate of exponential decay over time. These functions ensure that the exact solution respects the obstacle constraint \( u(x,t) \geq \psi(x,t) \).

The neural network is tasked with approximating the solution \( u(x,t) \) while minimizing the total loss function, which is composed of the PDE residual loss, obstacle penalty loss, boundary condition loss, and initial condition loss. The network is trained over 5000 iterations using a penalty parameter \( 1/\epsilon = 1e-5 \), which controls the strength of the obstacle constraint and enforces \( u(x,t) \geq \psi(x,t) \) via soft penalization.

During training, we use \( N_{\text{PDE}} = 1000 \) randomly sampled collocation points in the interior of the domain for the PDE residual and obstacle loss, \( N_{\text{boundary}} = 1000 \) points on the spatial boundary, and \( N_{\text{initial}} = 1000 \) points for the initial condition at \( t = 0 \). The interior and boundary collocation points are re-sampled randomly at each training iteration to promote generalization and robustness of the learned solution. This strategy avoids overfitting to specific point configurations and ensures better coverage of the domain over time.

\begin{figure}[H]
    \centering
    \includegraphics[width=\textwidth]{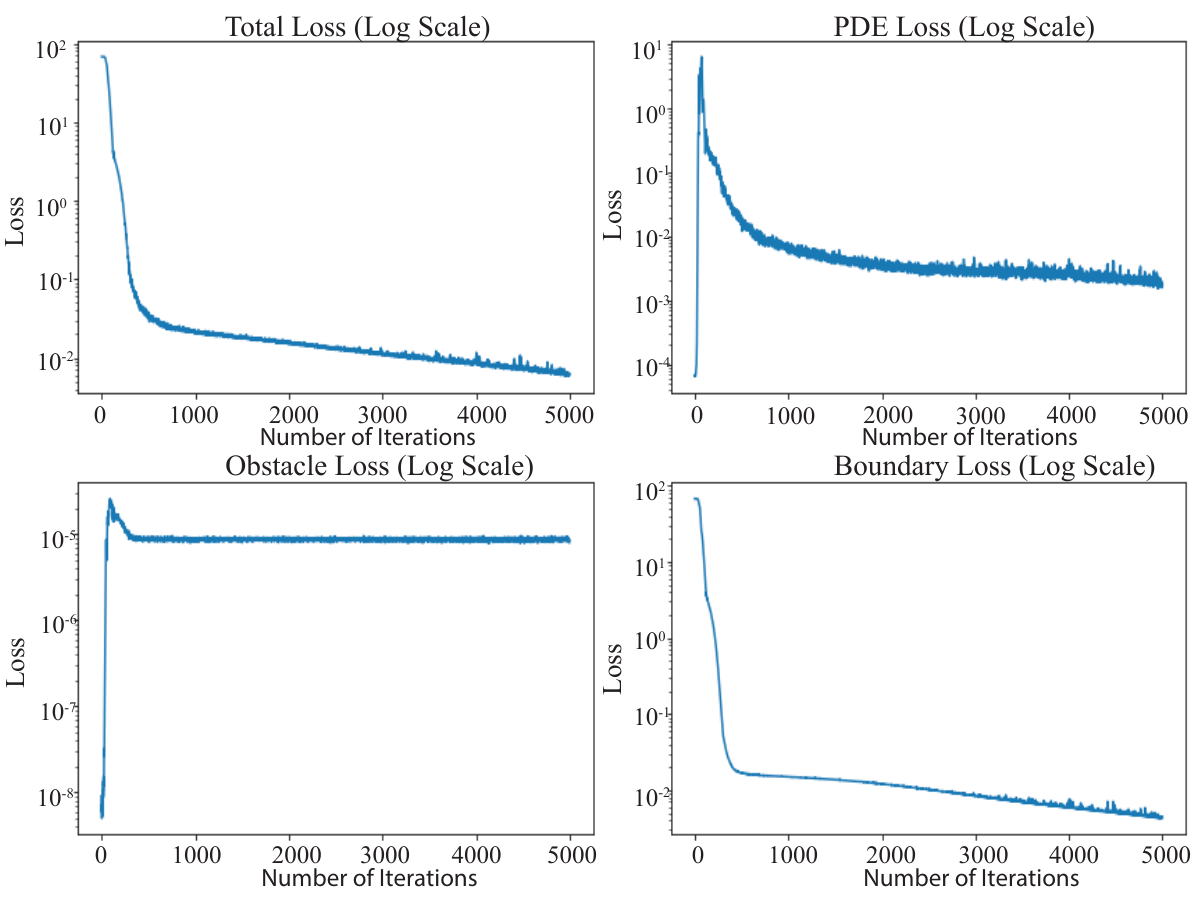}
    \caption{Loss components during training for the 1D parabolic obstacle problem. The figure shows the total loss, PDE loss, obstacle loss, and boundary condition loss.}
    \label{fig:losses1D}
\end{figure}

The evolution of the loss components during training is shown in Figure \ref{fig:losses1D}, where the total loss, PDE loss, obstacle loss, and boundary condition loss are displayed in a log scale, demonstrating effective convergence.Note that the initial condition loss was not plotted separately, as it closely overlapped with the boundary condition loss at $t=0$, and its values were negligible throughout training. Nevertheless, it was included in the total loss computation.

\begin{figure}[H]
    \centering
    \includegraphics[width=\textwidth]{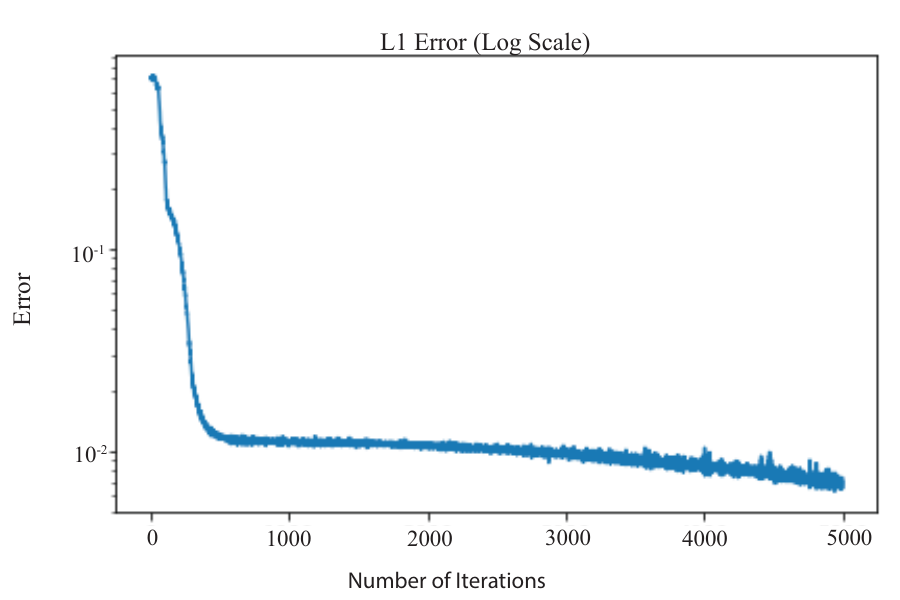}
    \caption{Evolution of the \( L^1 \) error for the 1D problem, showing the convergence of the neural network's approximation to the exact solution. The \( L^1 \) error is averaged over all sampled collocation points in both the spatial and temporal domains.}
    \label{fig:loss1D}
\end{figure}

Additionally, the evolution of the \( L^1 \) error during training is shown in Figure \ref{fig:loss1D}, where the error is averaged over the entire spatial and temporal domain. This demonstrates the convergence of the neural network's approximation to the exact solution across all collocation points.

\begin{figure}[H]
    \centering
    \includegraphics[width=\textwidth]{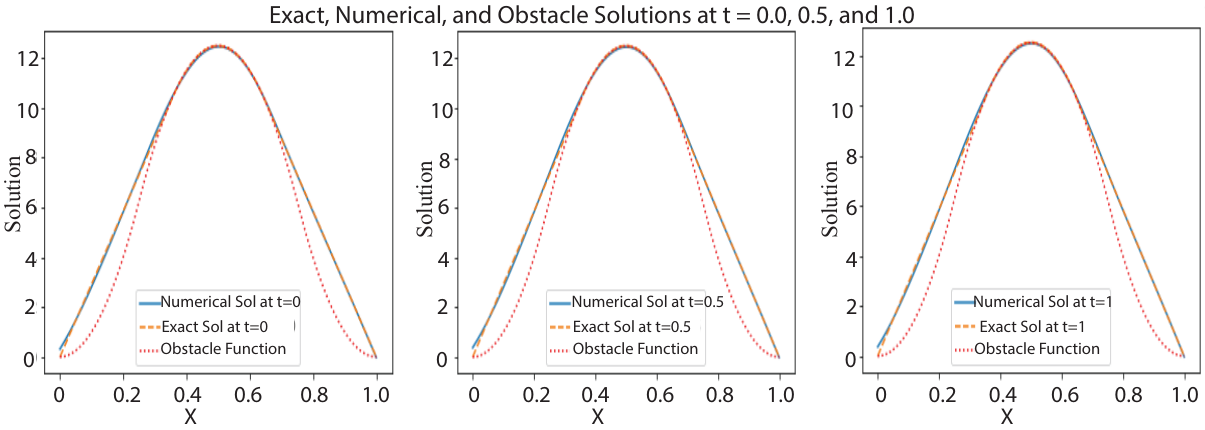}
    \caption{Comparison of neural network predictions with the exact and obstacle solutions for the 1D problem at \( t = 0.0 \), \( t = 0.5 \), and \( t = 1.0 \).}
    \label{fig:one_d_solutions}
\end{figure}

Finally, Figures \ref{fig:one_d_solutions} compare the neural network's prediction, the exact solution, and the obstacle at different time steps \( t = 0.0 \), \( t = 0.5 \), and \( t = 1.0 \) for the 1D problem. These comparisons illustrate the accuracy of the neural network in approximating the exact solution over time, while adhering to the obstacle constraint.

\subsection{Influence of Penalty Parameter \texorpdfstring{\( 1/\epsilon \)}{1/epsilon} on \texorpdfstring{$L^1$}{L1} Error}

The penalty parameter \( 1/\epsilon \), which governs the obstacle constraint, plays a significant role in controlling the accuracy of the solution. The parameter \( \epsilon \) is introduced in the obstacle constraint loss, and a smaller \( \epsilon \) leads to a stronger penalty, enforcing the obstacle condition more strictly.

\begin{table}[H]
\centering
\begin{tabular}{|c|c|}
\hline
\textbf{Penalty Parameter \( 1/\epsilon \)} & \textbf{$L^1$ Error} \\
\hline
$1e^{-1}$ & 0.40 \\
$1e^{-2}$ & 0.25 \\
$1e^{-4}$ & 0.35 \\
$1e^{-5}$ & 0.01 \\
$1e^{-8}$ & 0.33 \\
$1e^{-10}$ & 0.60 \\
\hline
\end{tabular}
\caption{$L^1$ errors corresponding to different values of the penalty parameter \( 1/\epsilon \).}
\label{tab:penalty_params}
\end{table}

Table \ref{tab:penalty_params} shows the relative errors for a range of values for \( 1/\epsilon \). This analysis shows that the penalty parameter \( 1/\epsilon \) plays a significant role in the accuracy of the model. When its value is too large, the obstacle condition is not sufficiently enforced. However, when it is too small, enforcement of the obstacle condition is prioritized at the expense of satisfying the PDE. Thus, a balance is required.

\section{2D Example: Comparative Study of Two Obstacle Functions}

In this section, we analyze a 2D parabolic obstacle problem over a radial spatial domain with a temporal domain \( t \in [0,1] \). Creating a non-trivial manufactured solution that temporally interacts with the obstacle is challenging. As such, we design two MMS test cases to examine the interaction between the analytical solution and the obstacle function.

1. \textbf{Case 1}: The obstacle function remains strictly below the analytical solution, allowing free evolution over time. In this case the solution changes significantly with time but there is little interaction between the solution and the obstacle.

2. \textbf{Case 2}: The obstacle function coincides with the analytical solution in specific regions, restricting movement. In this case the obstacle significantly influences the solution, but the temporal dynamics of that solution are more restricted.

The following subsections provide a detailed comparative analysis, including visualization, analytical formulation, and numerical approximation for both cases.

\subsection{Comparison of Obstacle Functions}

The obstacle functions for the two cases are defined as follows:

\paragraph{Case 1: No Contact Between Obstacle and Analytical Solution}
\[
\psi_1(r, t) = 
\begin{cases} 
\sqrt{1 - r^2} - 0.7, & r \leq r_{\star}, \\
-1, & r > r_{\star}.
\end{cases}
\]

\paragraph{Case 2: Partial Contact Between Obstacle and Analytical Solution}
\[
\psi_2(r, t) = 
\begin{cases} 
\sqrt{1 - r^2}, & r \leq r_{\star}, \\
-1, & r > r_{\star}.
\end{cases}
\]

\noindent The primary distinction lies in Case 2, where the obstacle function touches the analytical solution in some regions, limiting movement.

\begin{figure}[H]
    \centering
    \includegraphics[width=0.49\textwidth]{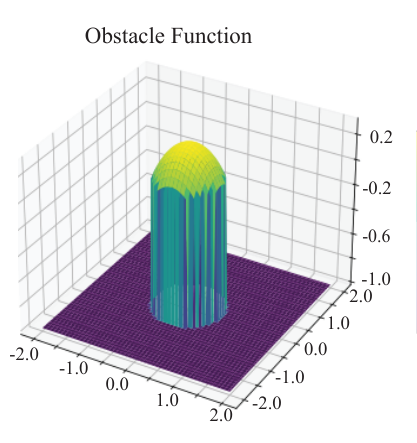}
    \includegraphics[width=0.49\textwidth]{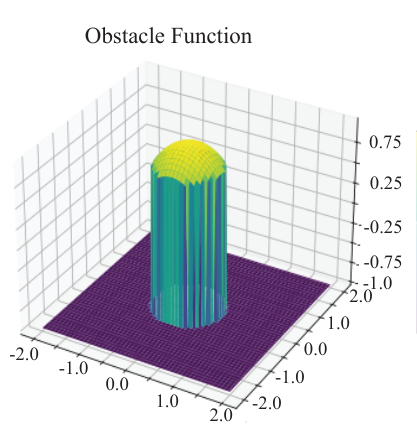}
    \caption{Obstacle functions for Case 1 (left) and Case 2 (right). In Case 1, the obstacle remains below the analytical solution, whereas in Case 2, it coincides with the analytical solution in some regions.}
    \label{fig:obstacle_comparison}
\end{figure}

\subsection{Comparison of Analytical Solutions}

The analytical solution for both cases is given by:

\[
    u(r, t) = 
    \begin{cases} 
    \sqrt{1 - r^2} \exp(- \gamma t), & r \leq r_{\star}, \\
    -\frac{r_{\star}^2 \log(r/2)}{\sqrt{1 - r_{\star}^2}} \exp(- \gamma t), & r > r_{\star}.
    \end{cases}
\]

\noindent where \( r_{\star} = 0.6979651482 \) and \( \gamma \) is a constant that is different for cases 1 and 2. This function evolves over time, constrained by the obstacle function in each case.

\subsection{Numerical Comparison of Analytical and Numerical Solutions}

The following figures compare the analytical and numerical solutions for both cases at different time steps (\( t = 0.0, 0.5, 1.0 \)).

\begin{figure}[H]
    \centering
    \includegraphics[width=\linewidth]{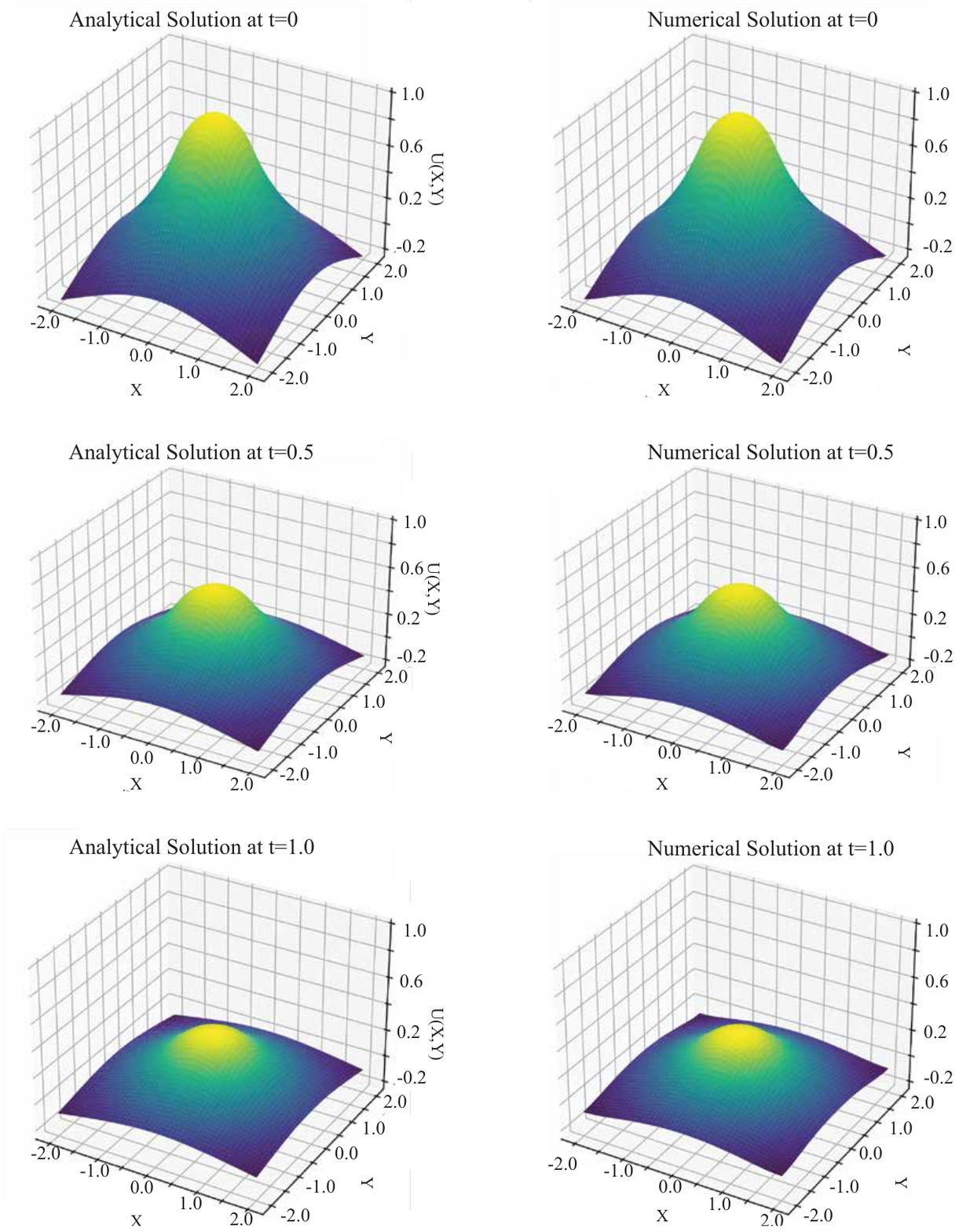}
    \caption{Comparison of analytical and numerical solutions for \textbf{Case 1} at different time steps. Top: \( t = 0.0 \). Middle: \( t = 0.5 \). Bottom: \( t = 1.0 \). Each row shows (from top to bottom) the analytical solution and the numerical solution for each time step.}
    \label{fig:par_soln_case_1}
\end{figure}

\begin{figure}[H]
    \centering
    \includegraphics[width=\linewidth]{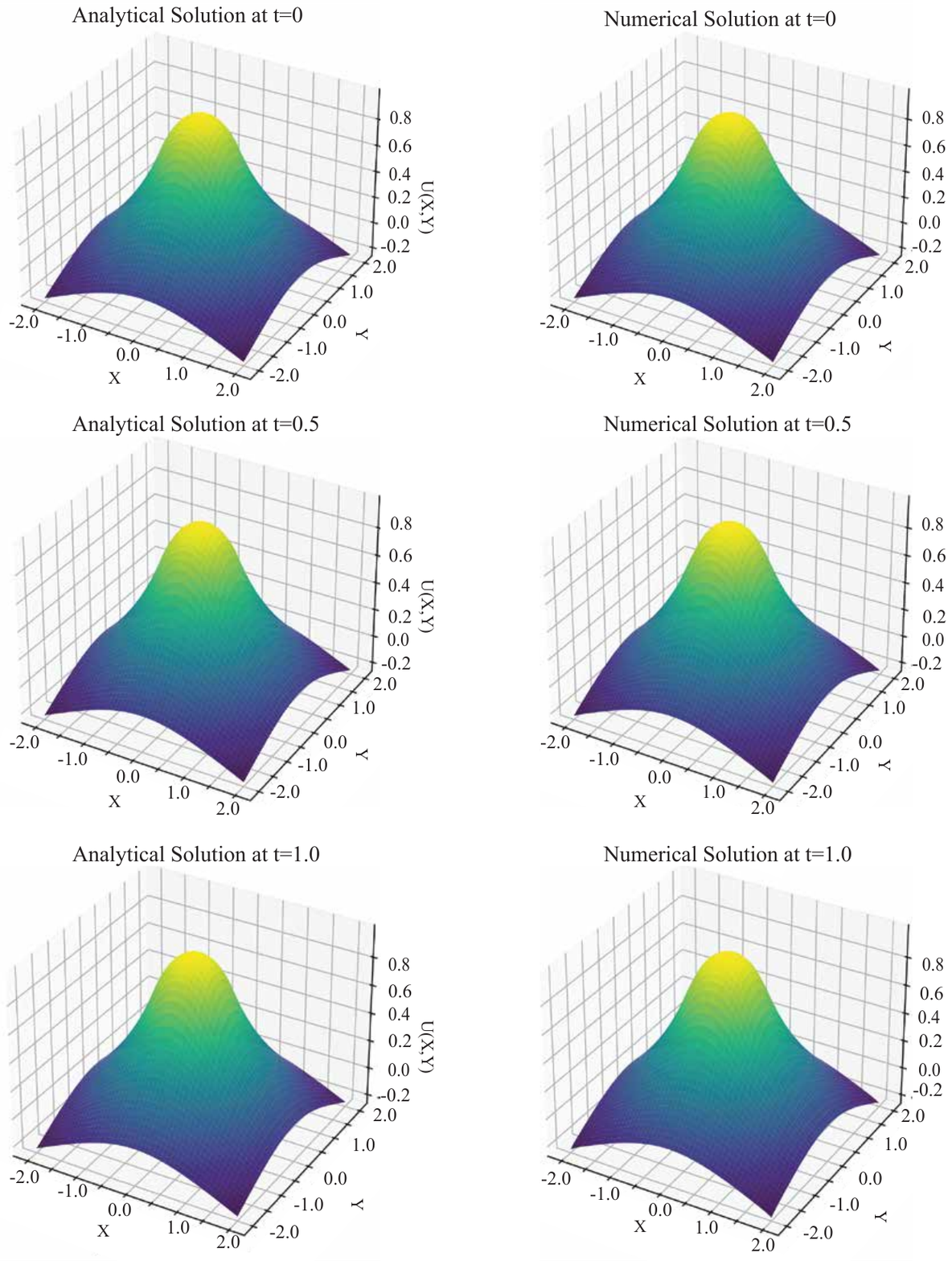}
    \caption{Comparison of analytical and numerical solutions for \textbf{Case 2} at different time steps. Top: \( t = 0.0 \). Middle: \( t = 0.5 \). Bottom: \( t = 1.0 \). Each row shows (from top to bottom) the analytical solution and the numerical solution for each time step. The obstacle function restricts movement in certain regions, impacting the solution's evolution.}
    \label{fig:par_soln_case_2}
\end{figure}

\subsection{Loss Components and \texorpdfstring{$L^1$}{L1} Error Analysis}
\begin{figure}[H]
    \centering
    \includegraphics[width=\textwidth]{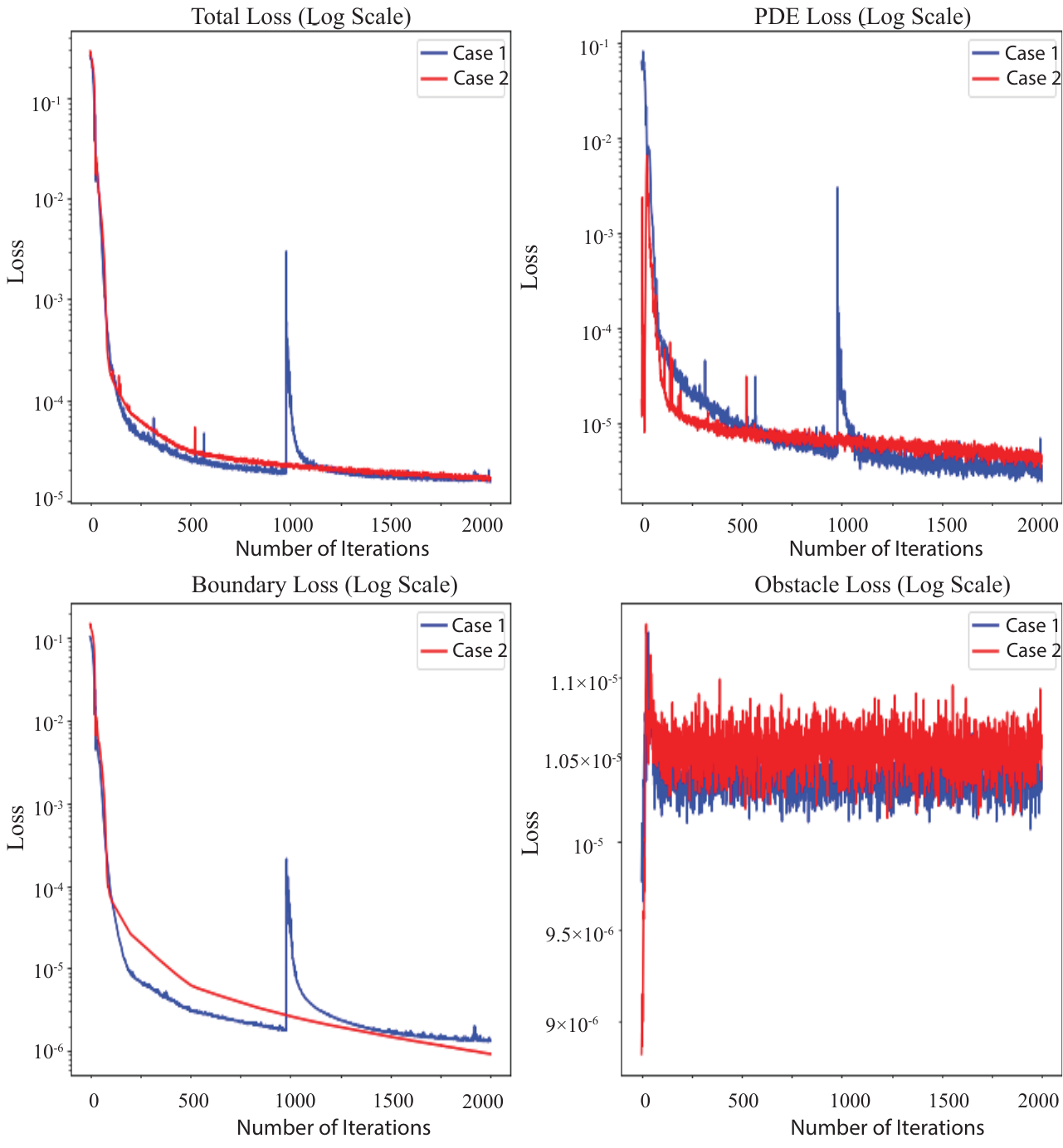}
    \caption{Loss components during training for the 2D problem. The figure presents the total loss, PDE loss, obstacle loss, and boundary condition loss in logarithmic scale over 2000 iterations. Different colors indicate Case 1 and Case 2, showing how the loss components evolve distinctly for each scenario.}
    \label{fig:all_losses_combined_2D}
\end{figure}

\begin{figure}[H]
    \centering
    \includegraphics[width=\textwidth]{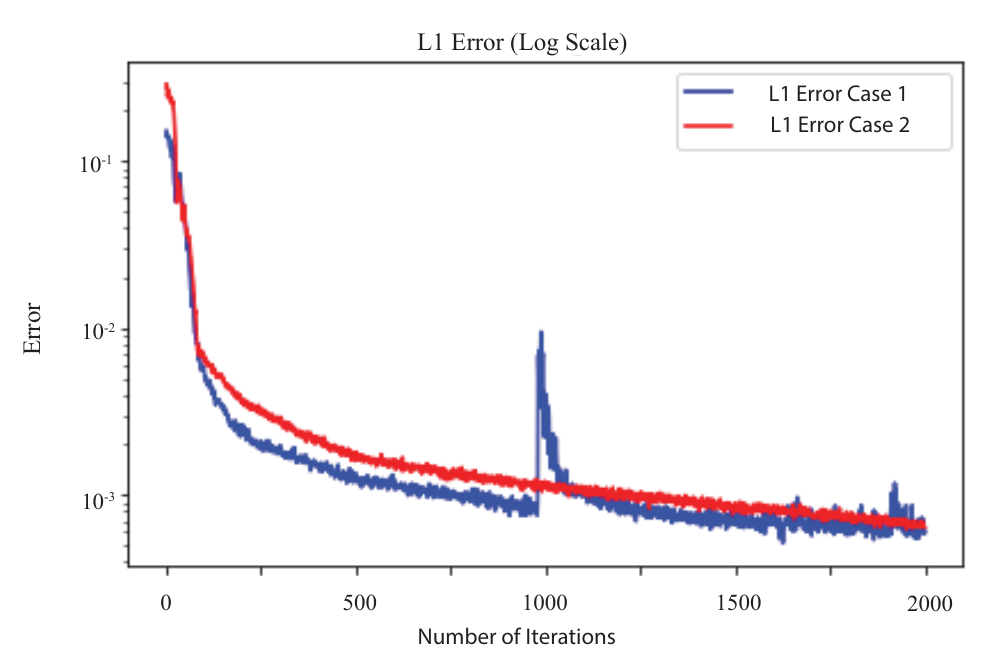}
    \caption{Evolution of the \( L^1 \) error for the 2D problem. The plot shows the convergence of the numerical solution towards the analytical solution, with separate curves for Case 1 and Case 2.}
    \label{fig:L1Error2D}
\end{figure}

The neural network minimizes the total loss, which includes PDE loss, obstacle loss, and boundary condition loss. Figure \ref{fig:all_losses_combined_2D} shows the evolution of these loss components over 2000 iterations for both Case 1 and Case 2, with different colors used to distinguish each case. The evolution of the \( L^1 \) error is presented in Figure \ref{fig:L1Error2D}, demonstrating the convergence of the numerical approximation to the analytical solution.

These results demonstrate that the neural network effectively learns the behavior of the parabolic obstacle problem under different constraints, validating the numerical approach.

\section{Mathematical Model of Ice Sheet Dynamics}\label{sec:model}

In this section, we present the mathematical formulation of the model, adapted from \cite{Ed2012, PT2023}, and describe the evolution of ice sheet dynamics. This model considers the surface elevation of the ice sheet relative to the underlying lithosphere and the evolution of ice thickness.

\subsection{Model Assumptions}
Before we proceed with the governing equations, we outline the key assumptions of the model:
\begin{itemize}
    \item The ice sheet's surface elevation \(h(t, x)\) is constrained to remain above the lithosphere elevation (the obstacle) \(b(x)\), for all \(x \in \Omega\) and for all times \(t \in [0, T]\).
    \item The ice thickness \(H(t, x) = h(t, x) - b(x)\) is constrained to be non-negative, resulting in an obstacle problem where the lithosphere elevation \(b(x)\) serves as the lower bound (obstacle) for the surface elevation.
    \item The basal sliding velocity \(U_b = 0\) when the base of the ice is frozen.
    \item The surface mass balance term \(a(t, x)\) depends on time, space, and the surface elevation \(h(t, x)\), while basal melting is neglected (\(a_b = 0\)).
\end{itemize}

\subsection{Governing Equations}

Let \( \Omega \) be a two-dimensional bounded domain in \( \mathbb{R}^2 \), representing the spatial extent of the ice sheet. For any point \( x = (x_1, x_2) \in \bar{\Omega} \), the surface elevation of the ice sheet is represented by \( h(t, x) \), where \(t \in [0, T]\).

The evolution of the ice sheet's surface elevation, \(h(t, x)\), is subject to the constraint that the surface remains above the lithosphere elevation, denoted by \(b(x)\):
\begin{equation}
h(t, x) \geq b(x), \quad \forall x \in \Omega, \, t \in [0, T].
\end{equation}
Here, \(b(x)\) describes the topography of the lithosphere, with \(b(x) > 0\) representing elevations above sea level and \(b(x) < 0\) representing those below.

The ice thickness \(H(t, x)\) is defined as the difference between the surface elevation and the lithosphere:
\begin{equation}
H(t, x) = h(t, x) - b(x), \quad H(t, x) \geq 0.
\end{equation}
This formulation introduces an obstacle problem where the obstacle is the lithosphere elevation \(b(x)\).

At each time \(t \in (0, T)\), the region covered by ice is defined as:
\begin{equation}
\Omega_t^+ = \{x \in \Omega \mid h(t,x) > b(x)\} = \{x \in \Omega \mid H(t,x) > 0\}.
\end{equation}
The corresponding free boundary \cite{Brezis1972}, denoted by \(\Gamma_{f,t}\), represents the boundary between the ice-covered region \(\Omega_t^+\) and the region devoid of ice \(\Omega_t^- = \Omega \setminus \Omega_t^+\). The free boundary is mathematically expressed as:
\begin{equation}
\Gamma_{f,t} = \Omega \cap \partial \Omega_t^+.
\end{equation}

The evolution of the ice sheet is governed by the conservation of mass, influenced by two primary sources: the surface mass balance, \(a_s\), and the basal melting rate, \(a_b\). When the basal temperature is below the melting point, we assume the basal melting rate is negligible (\(a_b = 0\)), so the dynamics focus on the surface mass balance term, \(a(t, x) = a_s(t, x, h)\).

The total ice flux, \(Q(t,x)\), which represents the horizontal volume of ice moving across the surface, is given by:
\begin{equation}
Q(t, x) = \int_b^h U(t, x, z) \, dz,
\end{equation}
where \(U(t, x, z)\) denotes the horizontal ice velocity at elevation \(z\), and the ice velocity is primarily driven by the surface gradient \(\nabla h\). Since the base is frozen \cite{Greve2009} (\(U_b = 0\)), the velocity field is determined purely by internal deformation of the ice.

We also assume that there is no ice flux across the free boundary \cite{Ed2012}:
\begin{equation}
Q \cdot \nu = 0 \quad \text{on} \quad \Gamma_{f,t},
\end{equation}
where \(\nu\) is the outward normal vector to the boundary. Moreover, since there is no ice in the region not covered by the ice sheet, we set:
\begin{equation}
Q = 0 \quad \text{in} \quad \Omega_t^-.
\end{equation}
Additionally, the ice thickness satisfies:
\begin{equation}
H = 0 \quad \text{on} \quad \Gamma_{f,t}, \quad \text{and} \quad H = 0 \quad \text{on}~ \partial \Omega.
\end{equation}

\subsection{Ice Thickness Equation}

The dynamics of ice thickness \(H(t, x)\) are governed by the balance between the flux divergence and the surface mass balance \cite{Greve2009}:
\begin{equation}
\frac{\partial H}{\partial t} = - \nabla \cdot Q + a(t, x).
\end{equation}
This equation captures how the ice thickness evolves in response to horizontal ice fluxes and surface accumulation or ablation.

\subsection{Variational Formulation}

To solve the ice thickness equation, we seek a variational formulation, which provides a more flexible framework for weak solutions, allowing us to handle the free boundary effectively. Let \(v = v(t,x)\) be any test function such that \(v(t, x) \geq b(x)\) for almost every \((t, x) \in (0, T) \times \Omega\). Multiplying the ice thickness equation by \((v - h)\) and integrating over \(\Omega\), we get:
\begin{equation}
\int_{\Omega} \frac{\partial H}{\partial t} (v - h) \, dx + \int_{\Omega} \nabla \cdot Q (v - h) \, dx = \int_{\Omega} a (v - h) \, dx.
\end{equation}

\noindent Applying the Gauss-Green theorem simplifies this to:
\begin{equation}
\int_{\Omega^+_t} \left( \frac{\partial H}{\partial t} + \nabla \cdot Q - a \right) (v - h) \, dx = 0 \quad \text{for all} \quad t \in (0, T).
\end{equation}

\noindent In regions where \(H = 0\), the equation becomes:
\begin{equation}
\frac{\partial H}{\partial t} \geq 0, \quad a \leq 0.
\end{equation}

\subsection{Boundary Value Problem}

The full boundary value problem is now described as finding \(H \geq 0\) such that:
\begin{equation}
\frac{\partial H}{\partial t} + \nabla \cdot Q - a = 0 \quad \text{in} \quad \Omega^+_t,
\end{equation}
and
\begin{equation}
\frac{\partial H}{\partial t} + \nabla \cdot Q - a \geq 0 \quad \text{in} \quad \Omega^-_t,
\end{equation}
with the boundary condition \(H(t, \cdot) = 0\) on \(\partial \Omega\) and the initial condition \(H(0, \cdot) = H_0 \geq 0\). Combining equations (2.7) and (2.14) leads to the following free boundary value problem:

\begin{equation}
\label{eq:free_boundary_value}
\left\{
\begin{aligned}
    \frac{\partial H}{\partial t} + \nabla \cdot Q - a &= 0, \quad \text{in} \quad \Omega^+_t, \\
    \frac{\partial H}{\partial t} + \nabla \cdot Q - a &\geq 0, \quad \text{in} \quad \Omega^-_t, \\
    H(t, \cdot) &= 0, \quad \text{on} \quad \partial \Omega, \\
    H(0, \cdot) &= H_0 \geq 0.
\end{aligned}
\right.
\end{equation}

\noindent The ice thickness dynamics described by equation \eqref{eq:free_boundary_value} outline the evolution of the ice sheet in both the ice-covered and ice-free regions.

\subsection{Ice Velocity and Flux}

To model the horizontal ice flow velocity \(U\), we rely on Glen's power law, which provides a relationship between the velocity and the surface elevation \(h\). The horizontal ice flow velocity \(U\) can be expressed as:
\begin{equation}
U( \cdot , \cdot , z) = -2( \rho g )^{p-1} \left( \int_z^b A(s)(h - s)^{p-1} \, ds \right) | \nabla h |^{p-2} \nabla h + U_b.
\end{equation}
This formulation assumes a frozen base where \(U_b = 0\) \cite{Greve2009}, and the velocity field is determined by internal deformation of the ice. Substituting this into the equation for ice flux \(Q\) gives:
\begin{equation}\label{q}
Q( \cdot , \cdot , z) = -2( \rho g )^{p-1} \left( \int_h^b \int_z^b A(s)(h - s)^{p-1} \, ds \, dz \right) | \nabla h |^{p-2} \nabla h + (h - b) U_b,
\end{equation}
where the second term represents basal sliding and the first term accounts for internal deformation. As the surface elevation \(h\) approaches the free boundary, the gradient of the surface norm tends to infinity, causing challenges in modeling near the boundary.

The viscosity of the ice is described in terms of the Glen power law with ice softness coefficient \( A(x,z) \) and exponent \( 2.8 \leq p \leq 5 \), with values for \( p \) suggested by laboratory experiments \cite{Goldsby2001}.

\subsection{Regularization of the Gradient}

To address the issue of gradient blow-up near the boundary, we introduce the following transformation, as proposed in \cite{Raviart1967}, which simplifies the analysis by regularizing the gradient near the boundary. Let:
\begin{equation}
H := u^{\frac{p-1}{2p}}.
\end{equation}

\noindent By using this transformation, \(H\) remains positive in \([0, T] \times \Omega\) if and only if \(u > 0\). Additionally, \(H(t,x) = 0\) if and only if \(u(t,x) = 0\), for all \(x \in \Omega\) and \(t \in [0, T]\). This transformation allows us to express the governing equations in a more tractable form.

Firstly, the time derivative of \(H\) becomes:
\begin{equation}
\frac{\partial H}{\partial t} = \frac{\partial}{\partial t} \left( |u|^{\frac{3p-1}{2p} - 2} u \right),
\end{equation}
which provides the rate of change of ice thickness in terms of \(u\). Secondly, the integral for ice softness \(A(s)\) is transformed as:
\begin{equation}
\int_b^h A(s)(h - s)^p \, ds = u^{\frac{(p+1)(p-1)}{2p}} \int_0^1 A(x, b + u^{\frac{p-1}{2p}} s')(1 - s')^p \, ds'.
\end{equation}
This simplifies the relationship between \(u\) and the ice softness. Thirdly, we transform the gradient term:
\begin{equation}
|\nabla h|^{p-2} \nabla h = \left( \frac{p - 1}{2p} \right)^{p-1} u^{-\frac{(p+1)(p-2)}{2p}} |\nabla u + u^{\frac{p+1}{2p}} \nabla b|^{p-2} u^{-\frac{p+1}{2p}} (\nabla u + u^{\frac{p+1}{2p}} \nabla b).
\end{equation}
This reformulation regularizes the gradient of the ice surface elevation \(h\) in terms of \(u\).

Now, for all \(x \in \Omega\) and for \(u = u(t, x) \in \mathbb{R}\), we define the vector field \(\Phi : \Omega \times \mathbb{R} \to \mathbb{R}^2\) by:
\begin{equation}
\Phi(x, u) := - \frac{2p}{p-1} u^{\frac{p+1}{2p}} \nabla b.
\end{equation}
Plugging \(\Phi\) into the previous equation for the gradient term, we get:
\begin{equation}
|\nabla h|^{p-2} \nabla h = \left( \frac{p-1}{2p} \right)^{p-1} u^{-\frac{(p+1)(p-2)}{2p}} |\nabla u - \Phi|^{p-2} (\nabla u - \Phi).
\end{equation}

For all \(x \in \Omega\) and for \(u \in \mathbb{R}\), we define another vector field \(\Psi : \Omega \times \mathbb{R} \to \mathbb{R}^2\) by:
\begin{equation}
\Psi(x, u) := (h - b) U_b.
\end{equation}
Additionally, we introduce the function \(\tilde{a} : [0, T] \times \Omega \times \mathbb{R} \to \mathbb{R}\), which modifies the surface mass balance term \(a(t, x)\) as:
\begin{equation}
\tilde{a}(t, x, u) := a(t, x, b + u^{\frac{p-1}{2p}}).
\end{equation}

Now, substituting the transformed expressions into the ice flux equation, we obtain:
\begin{equation}
\begin{split}
-Q &= 2 \left( \rho g \frac{p-1}{2p} \right)^{p-1} 
\left[ \int_0^1 A(x, b + u^{\frac{p-1}{2p}} s')(1 - s')^p \, ds' \right] \\
&\quad \times |\nabla u - \Phi|^{p-2} (\nabla u - \Phi) - \Psi.
\end{split}
\end{equation}

\noindent For simplicity, we define the function \(\mu : \Omega \times \mathbb{R} \to \mathbb{R}\) as:
\begin{equation}
\mu(x, u) := 2 \left( \rho g \frac{p-1}{2p} \right)^{p-1} \left[ \int_0^1 A(x, b + u^{\frac{p-1}{2p}} s')(1 - s')^p \, ds' \right].
\end{equation}

\noindent Substituting \(\mu(x, u)\) into the flux equation yields:
\begin{equation}
-Q = \mu(x, u) |\nabla u - \Phi|^{p-2} (\nabla u - \Phi) - \Psi.
\end{equation}

\noindent Finally, the ice thickness dynamics are described by the following unilateral boundary value problem:
\begin{equation}
\resizebox{\textwidth}{!}{$
\left\{
\begin{aligned}
    \frac{\partial}{\partial t} \left( |u|^{\frac{3p-1}{2p} - 2} u \right) - \nabla \cdot \left( \mu(x, u) |\nabla u - \Phi|^{p-2} (\nabla u - \Phi) - \Psi \right) &= \tilde{a}, \quad \text{in } \Omega^+_t, \\
    \frac{\partial}{\partial t} \left( |u|^{\frac{3p-1}{2p} - 2} u \right) - \nabla \cdot \left( \mu(x, u) |\nabla u - \Phi|^{p-2} (\nabla u - \Phi) - \Psi \right) &\geq \tilde{a}, \quad \text{in } \Omega^-_t, \\
    u(t, \cdot) &= 0, \quad \text{on } \partial \Omega, \\
    u(0, \cdot) &= u_0 := H_0^{\frac{2p}{p-1}} \geq 0.
\end{aligned}
\right.
$}
\end{equation}

\noindent The terms within this formulation capture the key dynamics of ice thickness evolution. However, they also introduce complexities due to the power-law dependence on \( u \), especially when \( u \) approaches zero.

\subsection{Singularity in the Time Derivative Term}

In this regularized model, the time derivative involves terms of the form:
\[
\frac{\partial}{\partial t} \left( |u|^{\frac{3p-1}{2p} - 2} u \right),
\]
where \( p \) lies within the range \( 2.8 \leq p \leq 5 \) \cite{Goldsby2001}. For these values of \( p \), this term can exhibit singularities when \( u \) becomes small. The power-law nature of this expression causes the time derivative to grow rapidly as \( u \) approaches zero, leading to potential numerical instability or non-physical oscillations in the solution. This sensitivity becomes particularly significant in physical regions where \( u \) is close to zero, making the equation challenging to solve accurately.

\subsubsection{Epsilon-Regularization to Manage the Singularity}

To address the singular behavior in the time derivative, we employ an \textit{epsilon-regularization} approach \cite{Evans2010}. This method ensures that the time derivative remains well-defined, even as \(u\) approaches zero. The regularized time derivative term is modified as follows:
\[
\frac{\partial}{\partial t} \left( \left( |u|^2 + \tilde{\epsilon} \right)^{\frac{3p-1}{4p} - 1} u \right),
\]
where \(\tilde{\epsilon} > 0\) is a small regularization parameter. This adjustment ensures that the term remains smooth and finite as \(u \to 0\), thus preventing abrupt changes or blow-ups in the solution. The regularization technique allows the model to maintain numerical stability while capturing the essential dynamics of the system for \(2.8 \leq p \leq 5\) \cite{Goldsby2001}.

With the mathematical model now stabilized through epsilon-regularization, we proceed to compute numerical solutions.

\section{Numerical Computation}

In this section, we use the method of manufactured to test the efficacy of this PINN-based approach for the 2D ice sheet problem with a known analytic solution to compare results against. The goal is to evaluate the accuracy and robustness of the proposed PINN-based solver using a known analytical solution that satisfies the governing PDE and obstacle constraint in Section 4.

We consider a radially symmetric, time-dependent manufactured solution defined as:
\[
u_{\text{exact}}(r, t) = u_0(r) \cdot e^{-\gamma t},
\]
where \( u_0(r) \) is the following spatial profile:
\[
u_0(r) = 
\begin{cases}
1 - \left( F(r) - G(r) + 1 - E(r) \right), & \text{for } r \leq r^*, \\
- \left(\frac{p}{p-1}\right) r^{*\frac{1}{p-1}} + (1 - r^*)^{(1/(p-1)) - 1} (r - r^*) \\
\quad + 1 - r^{*\frac{p}{p-1}} - (1 - r^*)^{\frac{p}{p-1}} + 1 - E(r^*), & \text{for } r > r^*,
\end{cases}
\]
with \( r = \sqrt{x^2 + y^2} \), \( r^* = 0.75 \), and the auxiliary functions defined as:
\[
F(r) = r^{\frac{p}{p-1}}, \quad G(r) = (1 - r)^{\frac{p}{p-1}}, \quad E(r) = \frac{p}{p-1} r.
\]

This solution is constructed to satisfy the obstacle condition \( u \geq b(r) \), where the obstacle function \( b(r) \) is given by:
\[
b(r) = 
\begin{cases}
1 - \left( r^{\frac{p}{p-1}} - (1 - r)^{\frac{p}{p-1}} + 1 - \frac{p}{p-1} r \right), & \text{for } r \leq r^*, \\
0, & \text{for } r > r^*.
\end{cases}
\]

\noindent The time decay \( e^{-\gamma t} \) introduces temporal dynamics while preserving the structure of the solution. The initial and boundary conditions are constructed to match this exact solution.

We use this setup to evaluate the performance of the PINN model for different values of the nonlinearity parameter \( p \). Accuracy is measured using the \( L^1 \) error between the predicted and exact solutions, and convergence is assessed via the total loss and its components.
The optimization procedure minimizes the total loss, which consists of the following components:

1. \textbf{PDE Residual Loss}: This term penalizes deviations from the governing ice thickness equation, ensuring that the neural network’s solution satisfies the PDE at a set of discrete points. The discrete version of the loss is defined as:
   \begin{small}
\begin{align}
\mathcal{L}_{\text{PDE}} &= \frac{1}{N} \sum_{i=1}^{N} \Bigg| 
\frac{\partial}{\partial t} \left( \left( |u(x_i, t_i)|^2 + \tilde{\epsilon} \right)^{\frac{3p-1}{4p} - 1} u(x_i, t_i) \right) \nonumber \\
&\quad - \nabla \cdot \left( \mu(x_i, u(x_i, t_i)) |\nabla u(x_i, t_i) - \Phi|^{p-2} (\nabla u(x_i, t_i) - \Phi) - \Psi \right) 
\Bigg|^2
\end{align}
\end{small}

\noindent where \( N \) is the total number of uniformly sampled discrete points \( (x_i, t_i) \) in the spatial-temporal domain \( \Omega \times (0, T) \), and \( \tilde{\epsilon} \) is a regularization parameter introduced to manage potential singularities.

2. \textbf{Obstacle Loss}: At discrete points, this loss becomes:
    \begin{equation}
        \mathcal{L}_{\text{obstacle}} = \frac{1}{N} \sum_{i=1}^{N} \left( \max(0, -u(x_i, t_i)) \right)^2,
    \end{equation}
    where \( N \) is the total number of sampled points \( (x_i, t_i) \) in the spatial domain \( \Omega \), ensuring the constraint \( u \geq 0 \) by penalizing values of \( u \) that fall below zero.

3. \textbf{Boundary Condition Loss}: For boundary points, this loss is computed as:
    \begin{equation}
    \mathcal{L}_{\text{boundary}} = \frac{1}{N_{\text{boundary}}} \sum_{i=1}^{N_{\text{boundary}}} \left( u(x_i, t_i) - g(x_i, t_i) \right)^2,
    \end{equation}
    where \( N_{\text{boundary}} \) is the total number of sampled points \( (x_i, t_i) \) on the boundary \( \partial \Omega \), enforcing \( u = 0 \) on the boundary.

4. \textbf{Initial Condition Loss}: This term enforces the correct initial profile \( u(x,0) = u_0(x) \) across the domain at time \( t = 0 \):
    \begin{equation}
    \mathcal{L}_{\text{initial}} = \frac{1}{N_{\text{initial}}} \sum_{i=1}^{N_{\text{initial}}} \left( u(x_i, 0) - u_0(x_i) \right)^2,
    \end{equation}
    where \( N_{\text{initial}} \) is the number of sampled points in the spatial domain \( \Omega \) at the initial time \( t = 0 \). This term ensures that the network prediction aligns with the known initial condition of the system.

The total loss is then defined as a weighted sum of these individual losses:
\[
\mathcal{L}_{\text{total}} = \alpha \mathcal{L}_{\text{PDE}} + \beta \mathcal{L}_{\text{obstacle}} + \gamma \mathcal{L}_{\text{boundary}} + \delta \mathcal{L}_{\text{initial}},
\]
where \( \alpha \), \( \beta \), \( \gamma \), and \( \delta \) are hyperparameters that control the relative importance of each loss term during training. In our experiments, we set \( \alpha = \gamma = \delta = 1.0 \) and \( \beta = 4000.0 \), giving equal weight to all components of the loss.

In this section, we demonstrate the performance of the method for different values of the parameter \( p \), which controls the nonlinearity in the ice thickness equation. Specifically, we present results for \( p = 2.8 \) and \( p = 4.0 \), focusing on the convergence of the total loss and the accuracy of the solution approximation.

\subsection{Results for \texorpdfstring {$p = 2.8$ }{p=2.8}}

\begin{figure}[H]
    \centering
    \includegraphics[width=\textwidth]{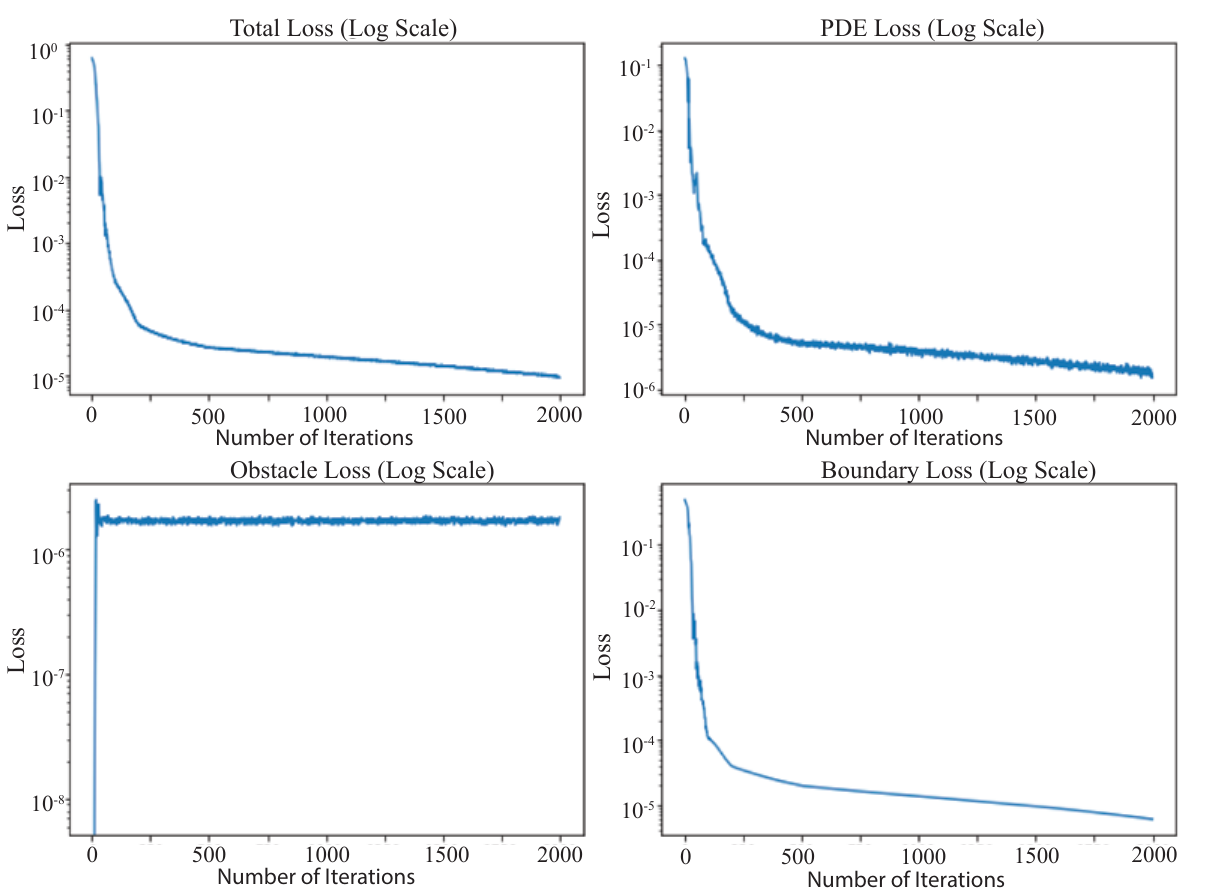}
    \caption{Loss components during training for \( p = 2.8 \). The total loss, PDE loss, obstacle loss and boundary condition loss are shown in log scale.}
    \label{fig:losses_28}
\end{figure}

Figure \ref{fig:losses_28} shows the evolution of the total loss, PDE loss, obstacle loss, and boundary condition loss during the training process for \( p = 2.8 \). The results are displayed on a log scale over 2000 iterations. The rapid decrease in the total loss and the individual loss components indicates that the network efficiently approximates the solution while satisfying the obstacle and boundary conditions.

\begin{figure}[H]
    \centering
    \includegraphics[width=0.9\textwidth]{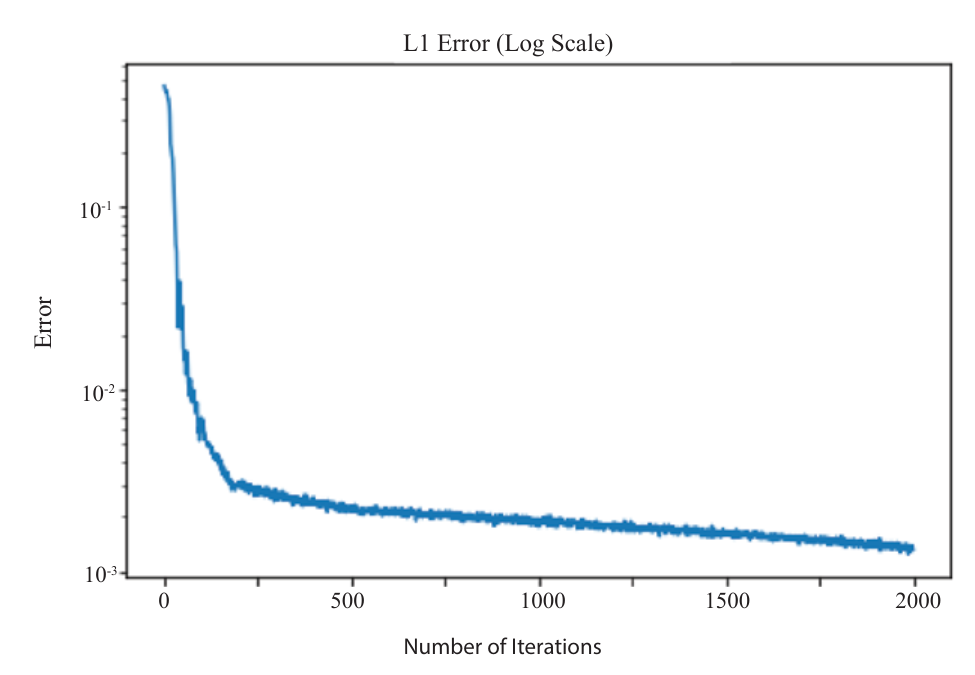}
    \caption{\( L^1 \) error during training for \( p = 2.8 \), shown in log scale.}
    \label{fig:L1Error_28}
\end{figure}

Figure \ref{fig:L1Error_28} presents the evolution of the \( L^1 \) error during training, which measures the discrepancy between the neural network's predicted solution and the exact solution. The \( L^1 \) error decreases rapidly during the initial iterations and continues to improve gradually, demonstrating the accuracy of the method.

\subsection{Results for \texorpdfstring {$p = 4.0$ }{p=4.0}}

\begin{figure}[H]
    \centering
    \includegraphics[width=\textwidth]{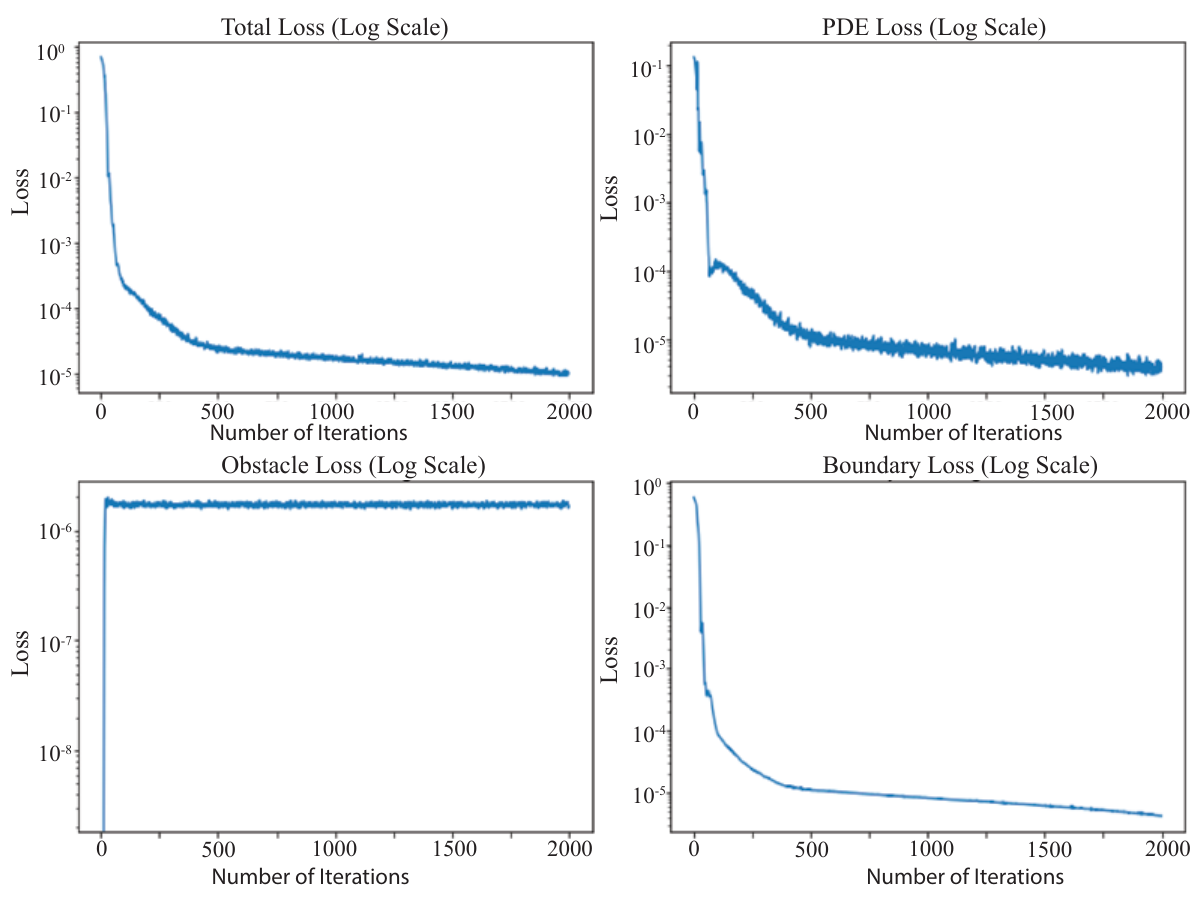}
    \caption{Loss components during training for \( p = 4.0 \). The total loss, PDE loss, obstacle loss, and boundary condition loss are shown in log scale.}
    \label{fig:losses_40}
\end{figure}

For \( p = 4.0 \), Figure \ref{fig:losses_40} illustrates the evolution of the total loss and the individual loss components over 2000 iterations. The loss curves are again shown in log scale. The network successfully minimizes the total loss, demonstrating stable convergence.

\begin{figure}[H]
    \centering
    \includegraphics[width=0.9\textwidth]{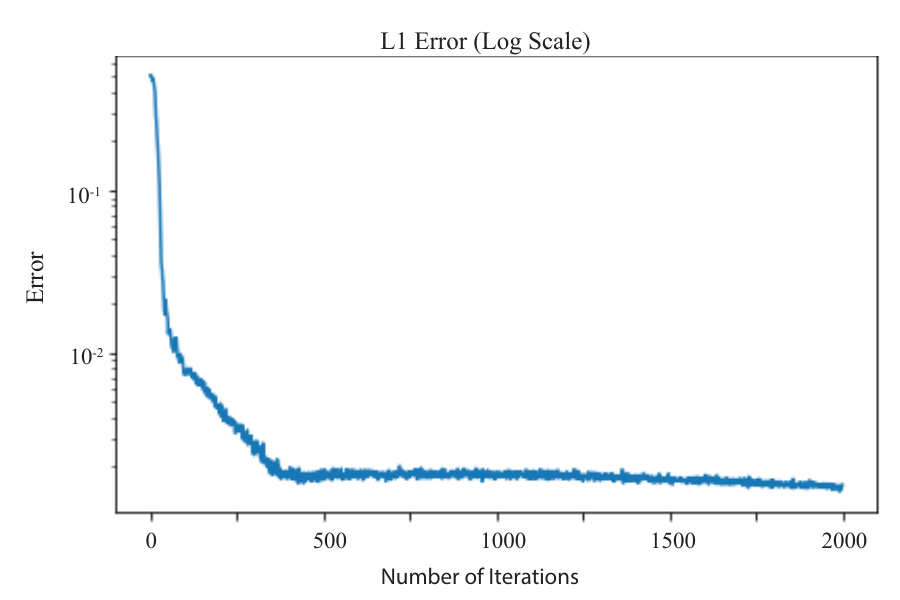}
    \caption{\( L^1 \) error during training for \( p = 4.0 \), shown in log scale.}
    \label{fig:L1Error_40}
\end{figure}

The evolution of the \( L^1 \) error for \( p = 4.0 \) is shown in Figure \ref{fig:L1Error_40}. The error decreases steadily, indicating that the network is learning an accurate approximation of the solution as the iterations progress.

\subsection{Solution Comparison for \texorpdfstring {$p = 4.0$ }{p=4.0}}

\begin{figure}[H]
    \centering
    \includegraphics[width=\textwidth]{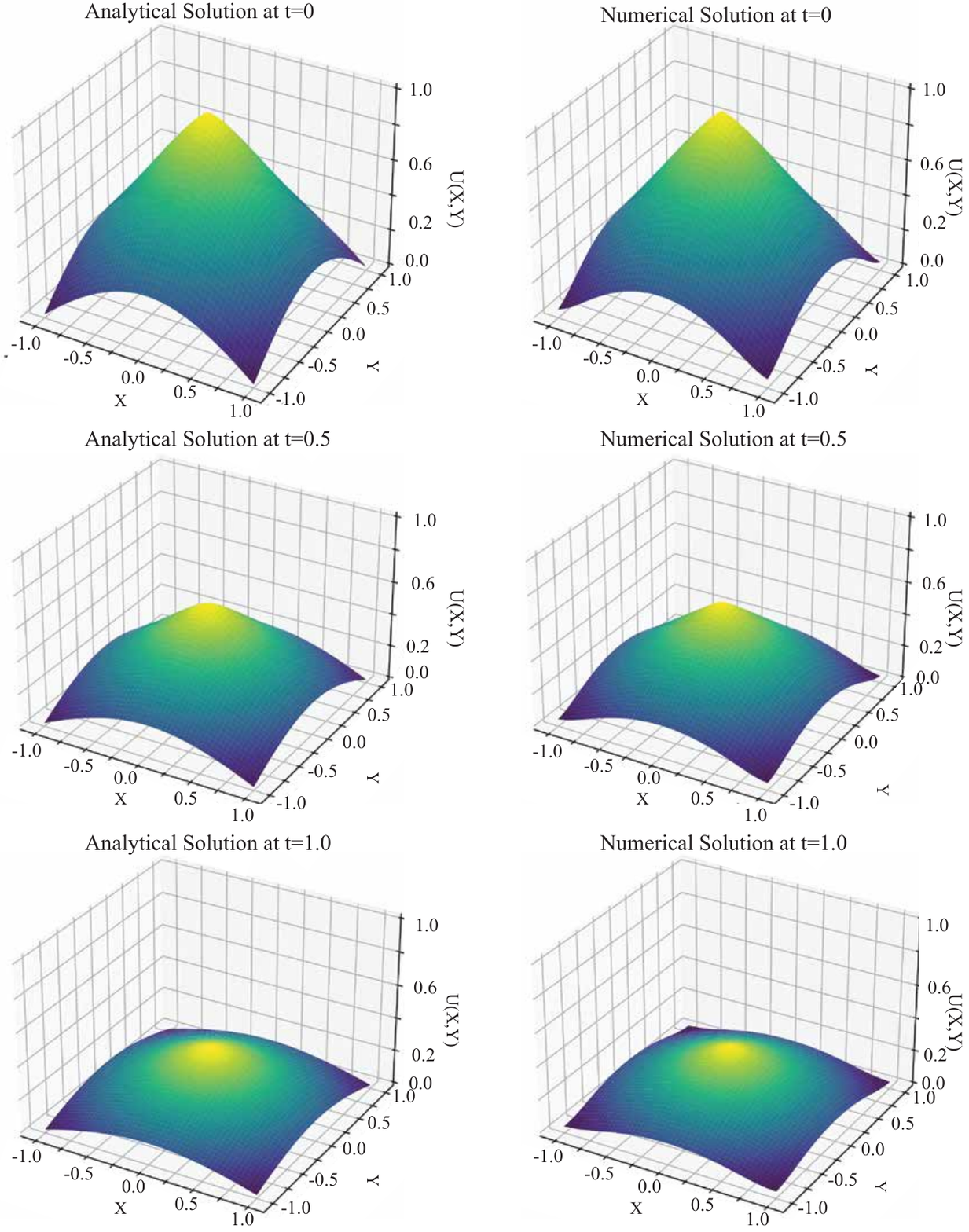}
    \caption{Comparison of analytical and numerical solutions at different time steps for \( p = 4.0 \). Top: \( t = 0.0 \), Middle: \( t = 0.5 \), Bottom: \( t = 1.0 \). Each row shows the analytical solution and the corresponding numerical approximation.}
    \label{fig:solutions_40}
\end{figure}

Figure \ref{fig:solutions_40} shows a comparison between the analytical solution and the numerical solution at three different time stamps: \( t = 0.0 \), \( t = 0.5 \), and \( t = 1.0 \), for \( p = 4.0 \). 

At \( t = 0.0 \), the numerical solution closely matches the analytical solution, indicating that the initial condition is well captured by the model. As the system evolves to \( t = 0.5 \), the numerical solution continues to align well with the analytical solution, demonstrating the model's ability to follow the time-dependent dynamics. By \( t = 1.0 \), the numerical solution remains in good agreement with the analytical one, reflecting the robustness of the method throughout the temporal evolution.

\section{Application to Devon Ice Cap Data}

\begin{figure}[H]
    \centering
    \includegraphics[width=0.8\textwidth]{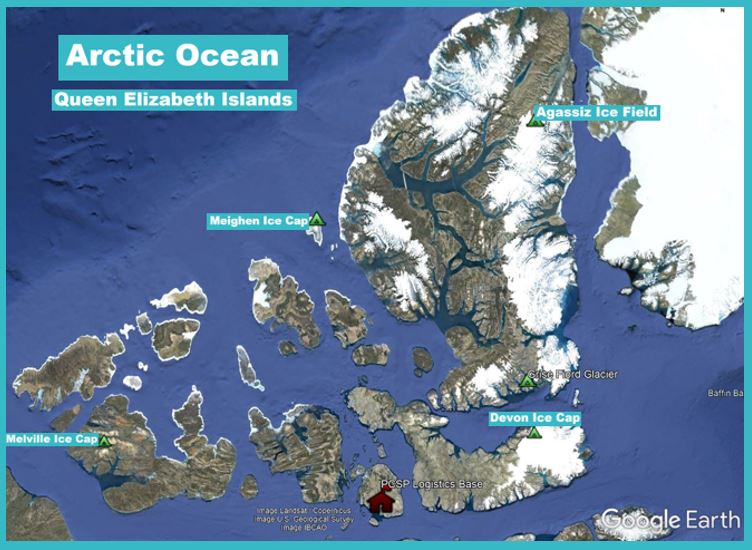}
    \caption{Map of the Canadian Arctic region showing the location of the Devon Ice Cap on Devon Island, along with other significant ice caps and glaciers. The map provides geographical context for the study area. Source: Natural Resources Canada, \url{https://natural-resources.canada.ca/simply-science/making-for-lost-time-canadas-far-north/24224}.}
    \label{fig:devon_ice_cap_map}
\end{figure}

In this section, we apply the techniques developed earlier to a dataset derived from the Devon Ice Cap in the Canadian Arctic. The Devon Ice Cap, located on Devon Island, is one of the largest ice caps in the region, covering an area of approximately 14,010 km². Figure \ref{fig:devon_ice_cap_map} shows a map of the Canadian Arctic, highlighting the location of the Devon Ice Cap.

The dataset used in this study includes ice thickness and surface elevation measurements from two different sources. The initial data comes from a 2000 airborne radar survey, which provided a comprehensive view of the ice cap's morphology \cite{Dowdeswell2004}. An updated aerogeophysical survey, conducted in 2018, offers new insights into changes in ice thickness and underlying topography over the 18-year period \cite{Rutishauser2022}. 

\subsection{Dataset Overview}

The Devon Ice Cap spans approximately 14,010 km², with the main ice cap covering 12,050 km². The initial data for this study comes from a year 2000 airborne radar survey, which recorded maximum ice thicknesses of up to 750 meters \cite{Dowdeswell2004}. In 2018, a follow-up aerogeophysical survey collected 4415 km of profile lines with line spacing ranging from 1.25 to 5 km \cite{Rutishauser2022}. We use the radar-derived digital elevation models (DEMs) to provide an initial condition for the ice thickness and the obstacle profile. We then train a PINN with this initial condition and obstacle to solve the PDE forward in time. Since $\mu$ is an unknown parameter in the model, we train multiple PINNs for  different values of $\mu$. These solution approximations are then used to propagate the initial condition forward in time for each value of $\mu$ and compare the predicted solution 18 years in the future to the 2018 data.

\subsection{Numerical Solution Consistency and Smooth Transition}

In this work, we developed a Physics-Informed Neural Network (PINN) model to solve the time-dependent PDEs governing ice thickness evolution. The 2000 ice thickness data \cite{Dowdeswell2004} serves as the problems initial condition, and the target is the 2018 ice thickness data \cite{Rutishauser2022}, obtained from aerogeophysical surveys. The network is trained to propagate the ice dynamics forward in time with the specific initial and bedrock data for the Devon Ice Cap.

In addition to ice thickness data, bedrock elevation data is incorporated as the obstacle condition for the model. This ensures that the predicted ice thickness does not fall below the physical constraints imposed by the underlying bedrock topography, providing a more realistic simulation of the ice sheet's interaction with the terrain.

\begin{figure}[H]
    \centering
    \includegraphics[width=0.45\textwidth]{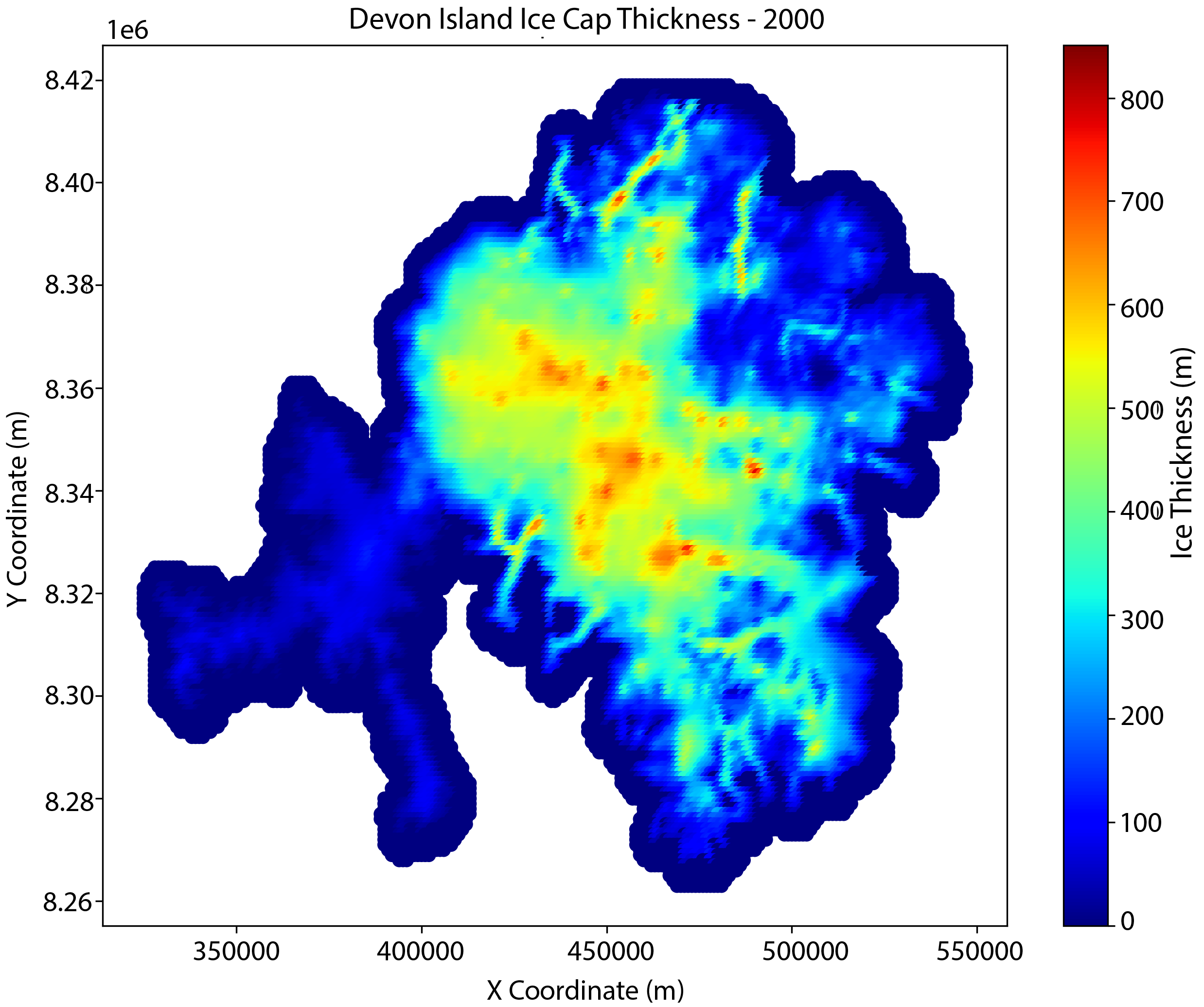}
    \includegraphics[width=0.45\textwidth]{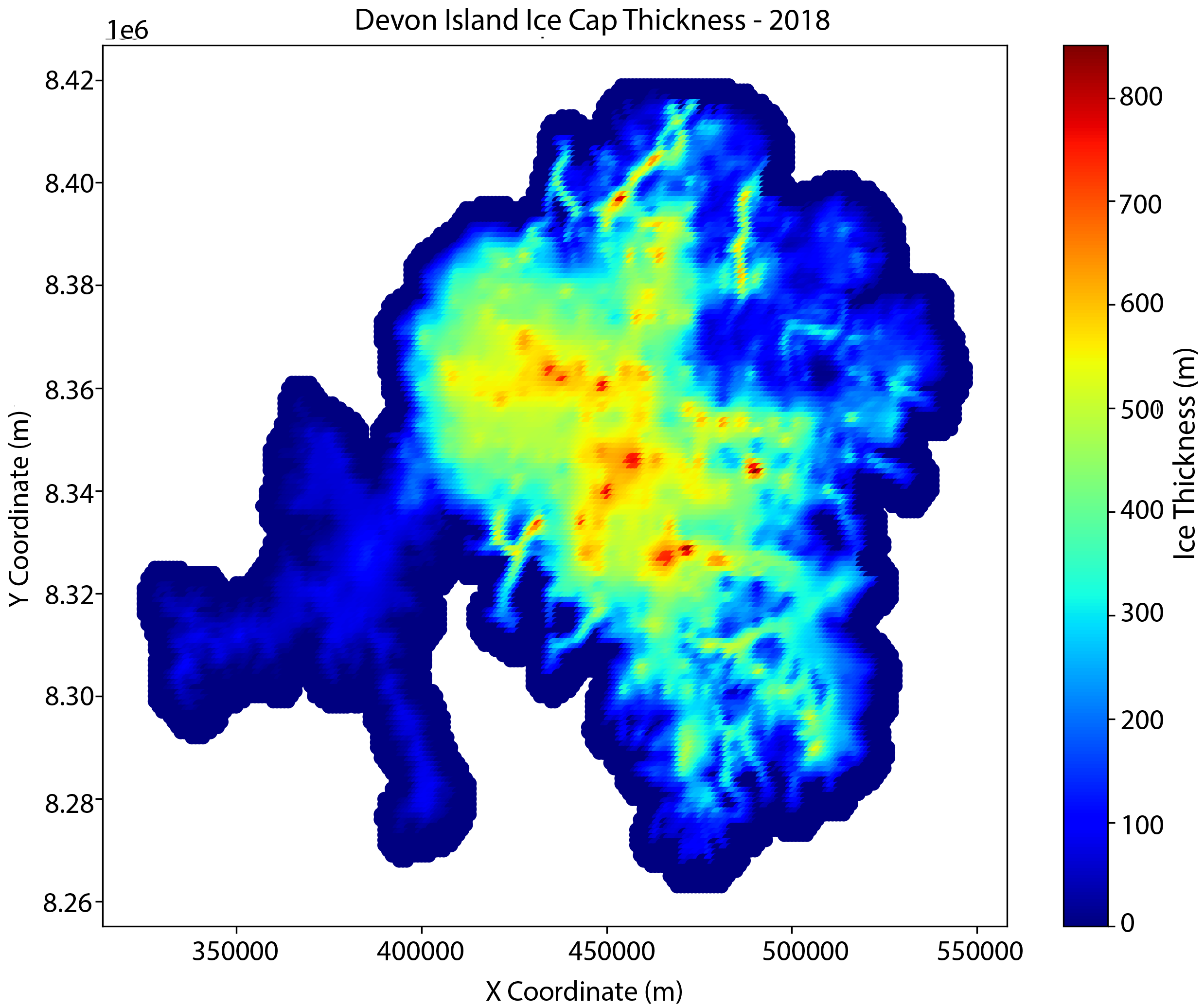}
    \caption{Comparison of Devon Ice Cap Ice Thickness: (Left) 2000 data used for initialization, (Right) 2018 data from aerogeophysical surveys. Both plots are scaled consistently for comparison.}
    \label{fig:ice_thickness_comparison}
\end{figure}

\begin{figure}[H]
    \centering
    \includegraphics[width=0.6\textwidth]{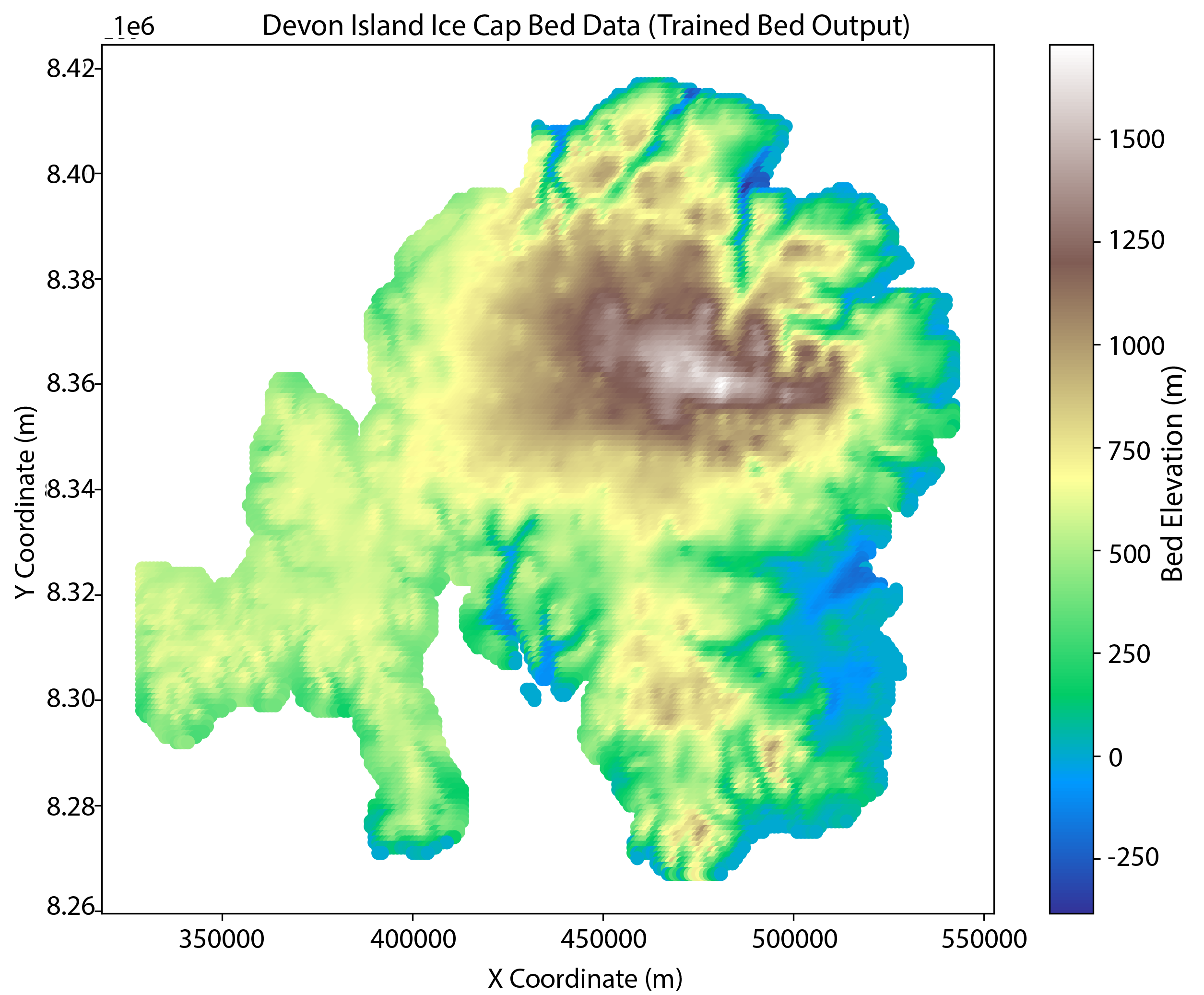}
    \caption{Bedrock elevation data for the Devon Ice Cap. This data is used as a constraint in the model, ensuring that the ice thickness evolution remains consistent with the physical boundaries of the bedrock.}
    \label{fig:bed_data}
\end{figure}

\subsection{Neural Network Initialization and Training}

Based on our prior work \cite{KC2024}, random initialization of parameters in a PINN to solve this type of problem with complex bedrock topography can lead to poor training performance. To address this, we utilize a two-step training process for our PINN:

1) \textbf{Initialization:} The network is initially trained to match the 2000 ice thickness data (\ref{fig:ice_thickness_comparison}). During this stage, the time is fixed at \(t = 0\), and the network learns \(u(x,y, 0)\), the ice thickness at the initial time. The same network architecture, which accepts \((x, y, t)\) as inputs, is used throughout both phases. During the initialization phase, \(t\) is fixed at \(t = 0\), effectively treating the problem as time-independent. This step allows the network to establish a stable and physically realistic initial solution. By doing so, we avoid the challenges associated with random initialization, such as erratic behavior and high error rates. Crucially, in this stage, the network is not solving the full time-dependent PDE; instead, it is focused on fitting the observed 2000 ice thickness data.

2) \textbf{Training for the Obstacle-PDE Problem:} Once the network has learned \(u(x,y,0)\), the network parameters are used as initial states for training the full time-dependent problem. The network then trains to solve the full obstacle-PDE, evolving the ice thickness over time based on the governing physics. This initialization ensures that the network begins the time-dependent training with a physically meaningful starting point, leading to faster convergence and reduced errors compared to random initialization.

\subsection{Training Process and Solution Approximation}

For the training process, we set the penalty parameters as \( \alpha = 1.0 \), \( \beta = 4000 \), \( \gamma = 1.0 \), and \( \delta = 1.0 \), determined by trial and error to balance the different components of the loss function. The relatively large value of \( \beta \) strongly enforces the obstacle constraint, while \( \alpha \), \( \gamma \), and \( \delta \) maintain the importance of the PDE residual, boundary, and initial condition losses, respectively.

The neural network architecture consisted of 15 hidden layers, each with 320 neurons, and was trained for 20,000 iterations to ensure accurate convergence. The training was conducted using the Adam optimizer with an initial learning rate of 
\(5 \times 10^{-3}\). A multi-step learning rate scheduler was employed to gradually decrease the learning rate during training, which helped stabilize convergence and improve the optimization of the loss components. After training, we use the resulting PINN to predict the ice thickness in 2018. Since \( \mu \) is a critical unknown parameter, we train PINNs to solve the PDE with different values of \( \mu \) and compare the predicted solution in 2018 to find the value that best captures the progression of the thickness over time.

\subsection{Analysis of Optimal \texorpdfstring{\(\mu\)}{mu} Values and Corresponding \texorpdfstring{$L^1$}{L1} Errors}

Table \ref{tab:mu_l1_error} illustrates how the values of the parameters \(\mu\) and \(p\) influence the level of agreement between the model's 2018 predictions and the 2018 data. Here, we quantify the \(L^1\) error (at the fixed time 2018) between the predicted solution and the observed data, using \(p = 2.8\) and \(p = 4\). From this analysis we identify the best values of $\mu$ and $p$ for future use (see the table caption).

\begin{table}[H] 
    \centering 
    \begin{tabular}{|c|c|c|} 
        \hline 
        \(\mu(x,u)\) & \(L^1\) Error (p = 2.8) & \(L^1\) Error (p = 4) \\ 
        \hline 
        \(5 \times 10^{-6}\)  & 0.0042 & 0.0031 \\ 
        \(5 \times 10^{-5}\)  & 0.0037 & 0.0024 \\ 
        \(5 \times 10^{-4}\)  & 0.00015 & 0.00001 \\ 
        0.005  & 0.0021 & 0.0015 \\ 
        0.05  & 0.0024 & 0.0019 \\ 
        0.5  & 0.0068 & 0.0051 \\ 
        1.0  & 0.014 & 0.011 \\ 
        2.0  & 0.53 & 0.41 \\ 
        \hline 
    \end{tabular} 
    \caption{Table of \(\mu\) values and their associated \(L^1\) errors for the 2018 Ice Thickness data at \(p = 2.8\) and \(p = 4\). The error trends help identify the optimal \(\mu\) and \(p\) combination. For subsequent analysis, we will use \(\mu = 5 \times 10^{-4}\) and \(p = 4\).}
    \label{tab:mu_l1_error} 
\end{table}

\subsection{Loss and Error Analysis for \texorpdfstring{\(\mathbf{p}=4\)}{p=4}}

\begin{figure}[H]
    \centering
    \includegraphics[width=\textwidth]{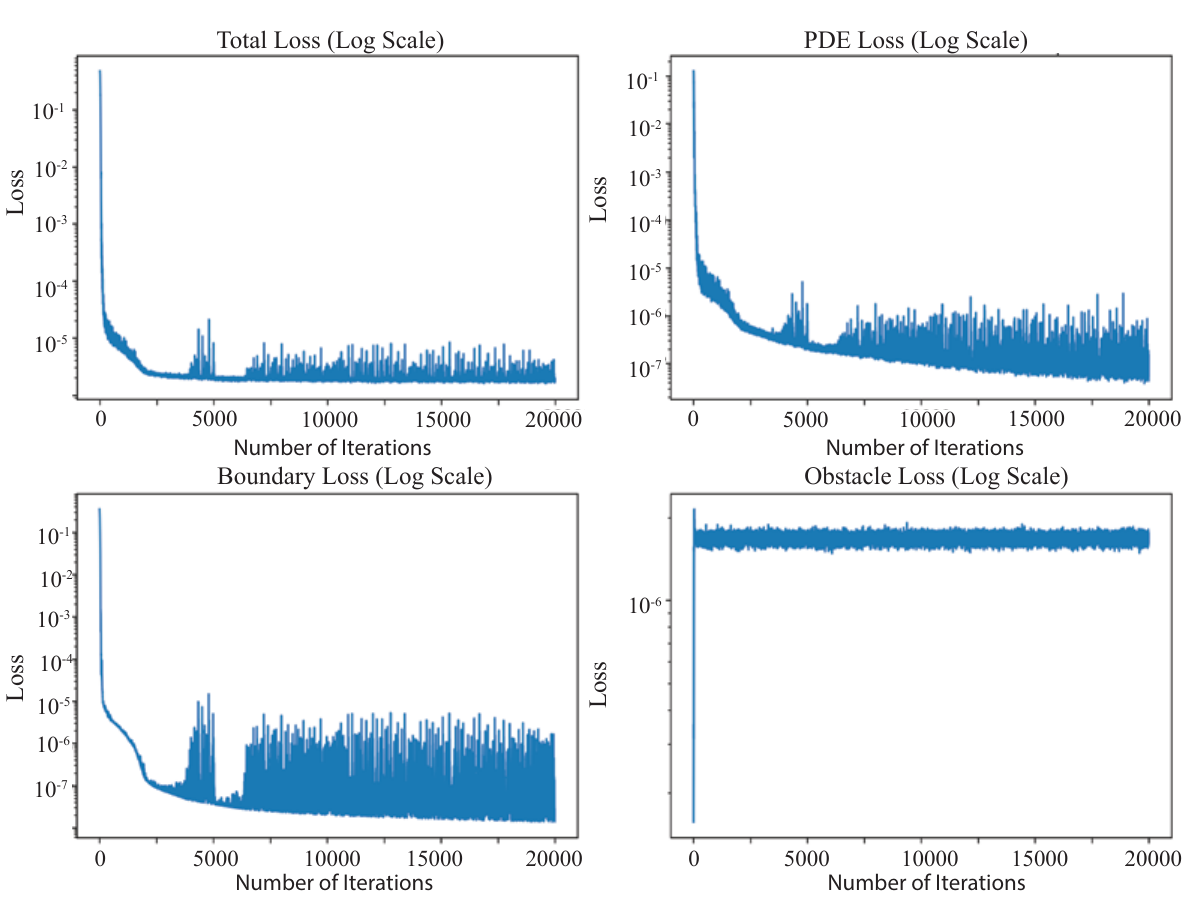}
    \caption{Loss Analysis for \(p = 4\), using the optimal value \(\mu = 5 \times 10^{-4}\) identified from Table~\ref{tab:mu_l1_error}. (a) Total Loss, (b) PDE Loss, (c) Boundary Condition Loss, and (d) Obstacle Loss over the training iterations (log scale).}
    \label{fig:loss_p4_combined}
\end{figure}

\begin{figure}[H]
    \centering
    \includegraphics[width=\textwidth]{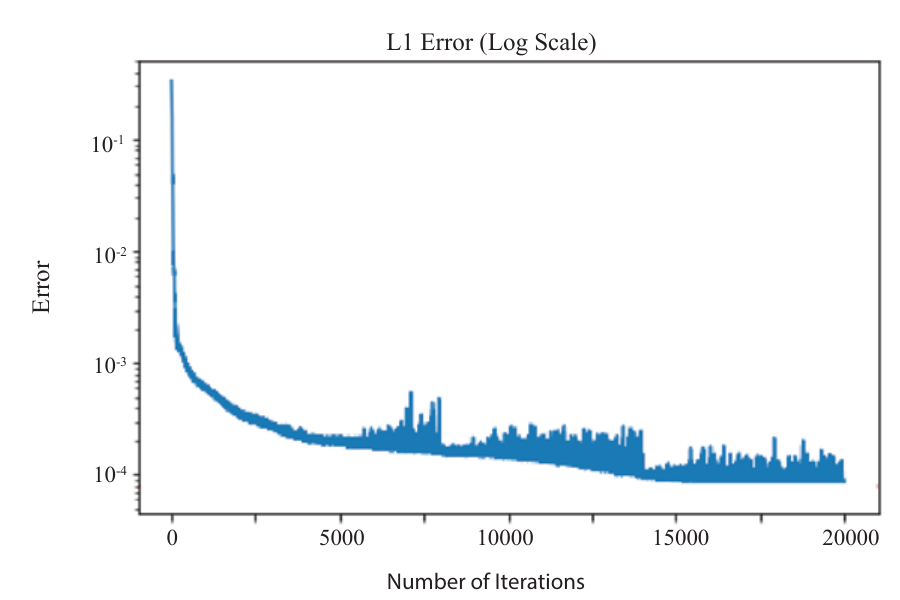}
   \caption{\(L^1\) Error Analysis for the 2018 Dataset. The plot illustrates the discrepancy between the model's predictions and the observed ice thickness data for 2018, using the optimal values \(\mu = 5 \cdot 10^{-4}\) and \(p = 4\), providing a measure of the model's accuracy at this specific time point.}
    \label{fig:l1_error_2018}
\end{figure}

Here we analyze the training metrics for the \( p=4 \) case. The loss components - total loss, PDE loss, obstacle loss, and boundary loss are illustrated in Figure \ref{fig:loss_p4_combined}. Figure \ref{fig:l1_error_2018} further illustrates the L1 error between computed solution and the data at 2018 as a function of training iteration. This metric provides a direct measure of the accuracy of the model at this time point, demonstrating the effectiveness of the model in capturing the observed ice thickness distribution at a future time point that is unseen during the training.

The \( L^1 \) error for the 2018 dataset, as shown in Figure \ref{fig:l1_error_2018}, reflects the precision of the model in predicting the observed ice thickness. This error metric, combined with the smooth convergence of all loss terms, further validates the robustness and precision of the model in capturing the dynamics of the Devon Ice Cap, particularly for the 2018 data.

\section{Conclusion}

This study presents a machine learning framework for solving parabolic obstacle problems in ice sheet modeling, with a focus on the Devon Ice Cap. By integrating traditional mathematical models with Physics-Informed Neural Networks (PINNs), our approach effectively captures the complex free-boundary conditions of grounded ice thickness dynamics. The framework was validated against data from 2000 and 2018, demonstrating its predictive capability for temporal changes in ice thickness and its scalability for different parameter settings.

Our results highlight the utility of PINNs in simulating ice sheet evolution under non-linear conditions, especially when data availability is limited. However, the model’s accuracy depends on parameter tuning and the quality of input data, such as the bedrock topography and surface elevation, which influences the network's ability to handle sharp gradients near the free boundary.

Future research could focus on enhancing the model by incorporating additional climate-related factors, such as temperature-dependent melting and basal sliding mechanisms. Additionally, applying this framework to other regions or ice caps would further validate its robustness and adaptability, potentially contributing to global climate models that forecast ice sheet contributions to sea level rise.

Overall, this study underscores the potential of machine learning approaches in geophysical modeling, bridging the gap between data-driven methods and traditional numerical modeling to address complex environmental challenges.
\section{Acknowledgments}

The authors acknowledge valuable discussions with Professor Ed Bueler, whose guidance was crucial in implementing the Shallow Ice Approximation (SIA) model on the Devon Ice Cap. This work was supported in part by the Research Fund of Indiana University.
\bibliographystyle{plainnat}
\bibliography{paper}
\newpage
\appendix
\section{Variational Inequality Formulation}
\label{appendix:variational_inequality}

For completeness, we present the variational inequality formulation associated with the ice thickness dynamics model. This formulation provides a framework to account for the free boundary dynamics of the ice-covered and ice-free regions.

The variational inequality formulation corresponding to the unilateral boundary value problem in Section \ref{sec:model} takes the following form: find \( u \geq 0 \) such that:
\[
\int_0^T \int_\Omega \left( \frac{\partial}{\partial t} \left( |u|^{\frac{3p-1}{2p} - 2} u \right) \right) (v - u) \, dx \, dt 
- \int_0^T \int_\Omega \mu(x, u) |\nabla u - \Phi|^{p-2} (\nabla u - \Phi) \cdot \nabla (v - u) \, dx \, dt 
\]
\[
\geq \int_0^T \int_\Omega \tilde{a}(t, x, u) (v - u) \, dx \, dt + \int_0^T \int_\Omega \Psi(x, u) \cdot \nabla (v - u) \, dx \, dt,
\]
for all test functions \( v \geq 0 \) such that \( v(t,x) \geq 0 \) for almost every \((t, x) \in (0, T) \times \Omega\).

The boundary conditions are:
\[
u(t, \cdot) = 0 \quad \text{on} \quad \partial \Omega, \quad \forall t \in (0, T),
\]
with the initial condition:
\[
u(0, \cdot) = u_0 = H_0^{\frac{2p}{p-1}} \geq 0.
\]

In this formulation, the time-dependent term captures the evolution of ice thickness \( u \), while the other terms account for internal deformation, advection, and basal sliding effects. This framework ensures that weak solutions can address the irregular boundary characteristics inherent in the ice sheet dynamics model.

\end{document}